\documentclass{article} 
\usepackage{iclr2021_conference,times}

\usepackage{amsmath,amsfonts,bm}

\def\eqref#1{equation~\ref{#1}}

\def\1{\bm{1}}

\DeclareMathAlphabet{\mathsfit}{\encodingdefault}{\sfdefault}{m}{sl}
\SetMathAlphabet{\mathsfit}{bold}{\encodingdefault}{\sfdefault}{bx}{n}

\usepackage{hyperref}
\usepackage{url}

\title{Graph Neural Networks with Learnable Structural and Positional Representations}

\author{%
  Vijay Prakash Dwivedi$^{1}$ \\
  \texttt{vijaypra001@e.ntu.edu.sg} \\
  \And
  Anh Tuan Luu$^1$ \\
  \texttt{anhtuan.luu@ntu.edu.sg} \\
  \And
  Thomas Laurent$^2$ \\
  \texttt{tlaurent@lmu.edu} \\
  \And
  Yoshua Bengio$^{3,4}$ \\
  \texttt{yoshua.bengio@mila.quebec} \\
  \And
  Xavier Bresson$^5$ \\
  \texttt{xavier@nus.edu.sg} \\
  \AND
  {\normalfont $^1$ Nanyang Technological University, Singapore \hspace{0.18cm} $^2$ Loyola Marymount University} \\
  $^3$ Mila, University of Montr\'eal \hspace{0.18cm} $^4$ CIFAR \hspace{0.14cm} $^5$ National University of Singapore
}

\usepackage{microtype}
\usepackage{graphicx}
\usepackage{booktabs} 

\usepackage{multirow}
\usepackage{url}
\usepackage{graphicx}
\usepackage{xcolor}
\usepackage{caption}
\usepackage{savesym}
\savesymbol{checkmark}
\usepackage{dingbat}
\usepackage{subcaption}
\usepackage{algorithm}

\usepackage{amsmath}
\usepackage{amssymb}

\newcommand{\best}[1]{{\color{red}#1}}

\usepackage{booktabs}
\usepackage{amssymb}
\usepackage{multirow}
\usepackage{sidenotes}
\usepackage{stmaryrd}

\iclrfinalcopy 
\begin{document}

\maketitle


\begin{abstract}
Graph neural networks (GNNs) have become the standard learning architectures for graphs.
GNNs have been applied to numerous domains ranging from quantum chemistry, recommender systems to knowledge graphs and natural language processing. A major issue with arbitrary graphs is the absence of canonical positional information of nodes, which decreases the representation power of GNNs to distinguish e.g. isomorphic nodes and other graph symmetries. An approach to tackle this issue is to introduce Positional Encoding (PE) of nodes, and inject it into the input layer, like in Transformers. Possible graph PE are Laplacian eigenvectors. In this work, we propose to decouple structural and positional representations to make easy for the network to learn these two essential properties. We introduce a novel generic architecture which we call \texttt{LSPE} (Learnable Structural and Positional Encodings). We investigate several sparse and fully-connected (Transformer-like) GNNs, and observe a performance increase for molecular datasets, from 
$1.79\%$
up to $64.14\%$
when considering learnable PE for both GNN classes.
\footnote{Code:  \url{https://github.com/vijaydwivedi75/gnn-lspe}}
\end{abstract}

\section{Introduction}
\label{sec:introduction}

GNNs have recently emerged as a powerful class of deep learning architectures to analyze datasets where information is present in the form of heteregeneous graphs that encode complex data connectivity. Experimentally, these architectures have 
shown great promises to be impactful in diverse domains such as drug design \citep{stokes2020deep, gaudelet2020utilising}, social networks \citep{monti2019fake, pal2020pinnersage}, traffic networks \citep{derrowpinion2021traffic}, physics \citep{cranmer2019learning, bapst2020unveiling}, combinatorial optimization \citep{bengio2021machine,cappart2021combinatorial} and medical diagnosis \citep{li2020graph}. 

Most GNNs (such as \cite{defferrard2016convolutional,sukhbaatar2016learning, kipf2017semi, hamilton2017inductive, monti2017geometric, bresson2017residual, velivckovic2018graph, xu2018how}) are designed with a message-passing mechanism \citep{gilmer2017neural} that builds node representation by aggregating local neighborhood information. It means that this class of GNNs is fundamentally structural, i.e. the node representation only depends on the  local structure of the graph. 
As such, two atoms in a molecule with the same neighborhood 
are expected to have similar representation. However, it can be limiting to have the same representation for these two atoms as their positions in the molecule are distinct, and their role may be specifically separate \citep{murphy2019relational}. As a consequence, the popular message-passing GNNs (MP-GNNs) fail to differentiate two nodes with the same 1-hop local structure. This restriction is now properly understood in the context of the equivalence of MP-GNNs with Weisfeiler-Leman (WL) test \citep{weisfeiler1968reduction} for graph isomorphism \citep{xu2018how, morris2019weisfeiler}. 

The said limitation can be alleviated, to certain extents, by (i) stacking multiple layers, (ii) applying higher-order GNNs, or (iii) considering positional encoding (PE) of nodes (and edges). Let us assume two structurally identical nodes in a graph with the same 1-hop neighborhood, but different with respect to 2-hop or higher-order neighborhoods. Then, stacking several layers \citep{bresson2017residual, li2019deepgcns} can propagate the information from a node to multiple hops, and thus differentiate the representation of two far-away nodes. However, this solution can be deficient for long-distance nodes because of the over-squashing phenomenon \citep{alon2020bottleneck}. Another approach is to compute higher-order node-tuple aggregations such as in WL-based GNNs \citep{maron2019provably, chen2019equivalence}; though these models are computationally more expensive to scale than MP-GNNs, even for medium-sized graphs \citep{dwivedi2020benchmarking}. An alternative technique is to consider a global  positioning of the nodes in the graph that can encode a graph-based distance between the nodes \citep{you2019position, dwivedi2020benchmarking,li2020distance,dwivedi2021generalization}, or can inform about specific sub-structures \citep{bouritsas2020improving,bodnar2021weisfeiler}.

\begin{figure}[!t]
    \centering
    \includegraphics[width=0.95\textwidth]{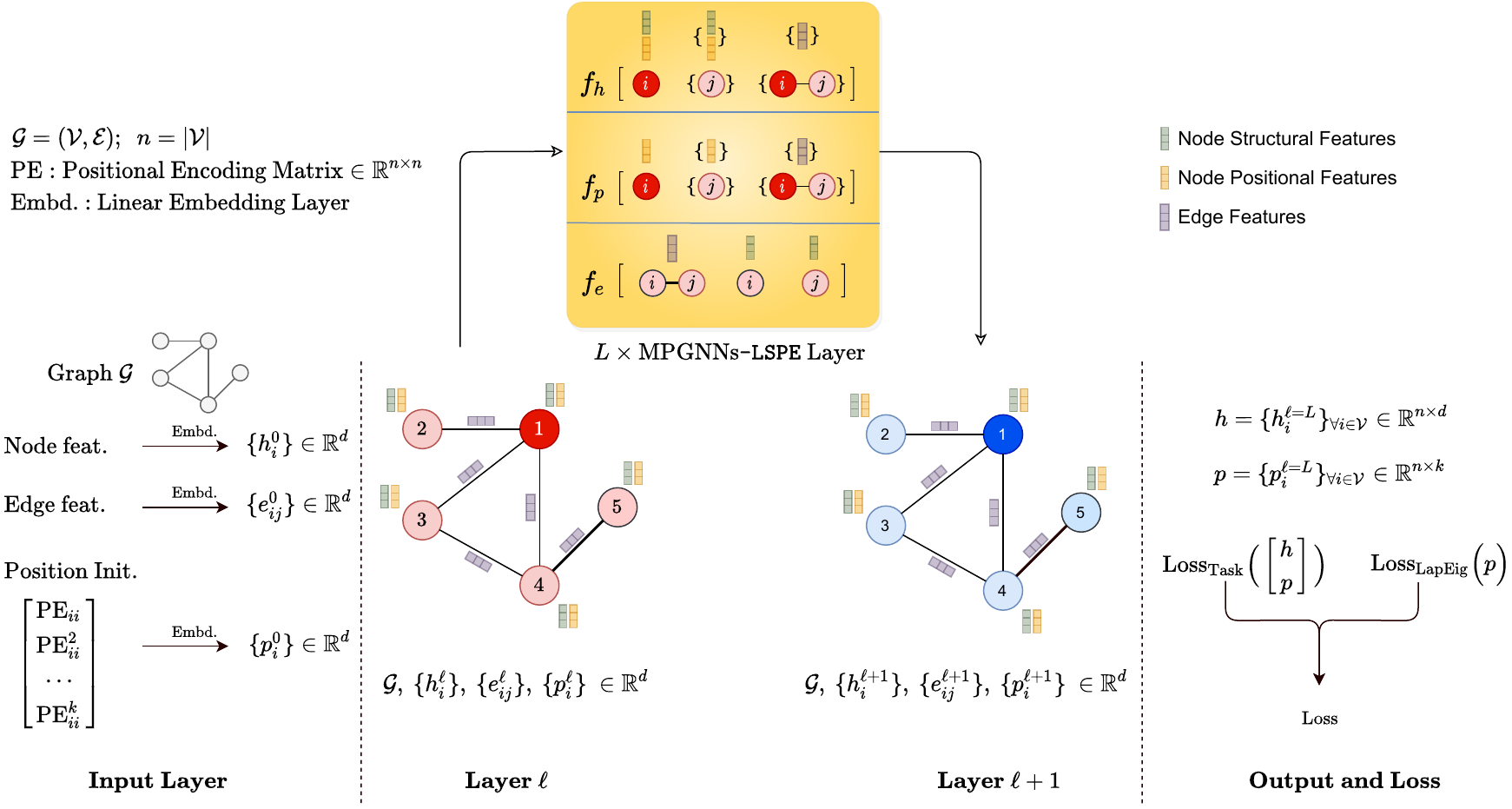}
    \vspace{-3pt}
    \caption{Block diagram illustration of the proposed MPGNNs\texttt{-LSPE} architecture along with the inputs, general framework of a layer, and the output and loss components.}
    \label{fig:mpgnns_lspe_architecture}
    \vspace{-9pt}
\end{figure}

{\bf Contribution.} In this work, we turn to the idea of {\it learning positional representation} that can be combined with structural GNNs to generate more expressive node embedding. Our main intent is to alleviate the lack of canonical positioning of nodes in arbitrary graphs to improve the representation power of MP-GNNs, while keeping their linear complexity for large-scale applications.
For this objective, we propose a novel framework, illustrated with Figure \ref{fig:mpgnns_lspe_architecture}, that enables GNNs to learn both structural and positional representations at the same time (thus named MPGNNs\texttt{-LSPE}).
Alongside, we present a random-walk diffusion based positional encoding scheme to initialize the positional representations of the nodes.
We show that the proposed architecture with learnable PE can be used with any graph network that fits to the MP-GNNs framework, 
and improves its performance 
($1.79\%$
to $64.14\%$).
In our demonstrations, we formulate \texttt{LSPE} instances of both sparse GNNs, such as GatedGCNs \citep{bresson2017residual} and PNA \citep{corso2020principal} and fully-connected Transformers-based GNNs \citep{kreuzer2021rethinking, mialon2021graphit}.
Our numerical experiments on three standard molecular benchmarks show that different instantiations of MP-GNNs with \texttt{LSPE} surpass the previous state-of-the-art (SOTA) on one dataset by a considerable margin 
($26.23\%$), while achieving SOTA-comparable score on 
the other two datasets.
The architecture also shows consistent improvements on three non-molecular benchmarks. In addition, our evaluations find the sparse MP-GNNs to be outperforming fully-connected GNNs, hence suggesting greater potential towards the development of highly efficient, yet powerful architectures for graphs.

\section{Related work}
\label{sec:related_work}

In this section, we review briefly the three research directions theoretical expressivity of GNNs, graph positional encoding, and Transformer-based GNNs.

\textbf{Theoretical expressivity and Weisfeiler-Leman GNNs.}
As the theoretical expressiveness of MP-GNNs is bounded by the 1-WL test \citep{xu2018how, morris2019weisfeiler}, they may perform poorly on graphs that exhibit several symmetries \citep{murphy2019relational}, and additionally some message-passing functions may not be discriminative enough \citep{corso2020principal}. To this end, $k$-order Equivariant-GNNs were introduced in \cite{maron2018invariant} requiring $O(n^k)$ memory and speed complexities. Although the complexity was improved to $O(n^2)$ memory and $O(n^3)$ respectively \citep{maron2019provably,chen2019equivalence,azizian2020expressive}, it is still inefficient compared with the linear complexity of MP-GNNs.

{\bf Graph Positional Encoding.}
The idea of positional encoding, i.e. the notion of global position of pixels in images, words in texts and nodes in graphs, plays a central role in the effectiveness of the most prominent neural networks with ConvNets \citep{lecun1998gradient}, RNNs \citep{hochreiter1997long}, and Transformers \citep{vaswani2017attention}. For GNNs, the position of nodes is more challenging due to the fact that there does not exist a canonical positioning of nodes in arbitrary graphs. Despite these issues, graph positional encoding are as much critical for GNNs as they are for ConvNets, RNNs and Transformers, as demonstrated for prediction tasks on graphs \citep{srinivasan2019equivalence, cui2021positional}. 
Nodes in a graph can be assigned index positional encoding (PE). However, such a model must be trained with the $n!$ possible index permutations or else sampling needs to be done \citep{murphy2019relational}. Another PE candidate for graphs can be Laplacian Eigenvectors \citep{dwivedi2020benchmarking, dwivedi2021generalization} as they form a meaningful local coordinate system, while preserving the global graph structure. However, there exists sign ambiguity in such PE as eigenvectors are defined up to $\pm1$, leading to $2^k$ number of possible sign values when selecting $k$ eigenvectors which a network needs to learn. Similarly, the eigenvectors may be unstable due to eigenvalue multiplicities. \cite{you2019position} proposed learnable position-aware embeddings based on random anchor sets of nodes, where the random selection of anchors has its limitations, which makes their approach less generalizable on inductive tasks. There also exists methods that encode prior information about a class of graphs of interest such as rings for molecules \citep{bouritsas2020improving, bodnar2021weisfeiler} which make MP-GNNs more expressive. But the prior information regarding graph sub-structures needs to be pre-computed, and sub-graph matching and counting require $O(n^k)$ for $k$-tuple sub-structure.

{\bf Transformer-based GNNs.}
Although sparse MP-GNNs are very efficient, they are susceptible to the information bottleneck limitation \citep{alon2020bottleneck} in addition to vanishing gradient (similar to RNNs) on tasks when long-range interactions between far away nodes are critical. To overcome these limitations, there have been recent works that generalize Transformers to graphs \citep{dwivedi2021generalization, kreuzer2021rethinking, ying2021transformers, mialon2021graphit} which alleviates the long-range issue as `everything is connected to everything'. However, these methods either use non-learnable PEs to encode graph structure information \citep{dwivedi2021generalization, ying2021transformers, mialon2021graphit}, or inject learned PEs to the Transformer network that relies on Laplacian eigenvectors \citep{kreuzer2021rethinking}, thus inheriting the sign ambiguity limitation.

A detailed review of the above research directions is available in the supplementary Section \ref{sec:related_work_detailed}. We attempt to address some of the major limitations of GNNs by proposing a novel architecture with consistent performance gains.

\section{Proposed Architecture}
\label{sec:proposed_architecture}

In this work, we decouple structural and positional representations to make it easy for the network to learn these two critical characteristics. This is in contrast with most existing architectures s.a. \cite{dwivedi2021generalization,beani2021directional,kreuzer2021rethinking} that inject the positional information into the input layer of the GNNs, and \cite{you2019position} that rely on distance-measured anchor sets of nodes limiting general, inductive usage.
Given the recent theoretical results on the importance of informative graph PE for expressive GNNs \citep{murphy2019relational,srinivasan2019equivalence,Loukas2020What},
we are interested in a generic framework that can enable GNNs to separate positional and structural representations to increase their expressivity.
Section \ref{sec:generic_formulation} will introduce our approach to  augment GNNs with learnable graph PE. Our framework can be used with different GNN architectures. We illustrate this flexibility in Sections \ref{sec:sparse_gnns_lspe} and \ref{sec:transformer_lspe} where the decoupling of structural and positional information is applied to both sparse MP-GNNs and fully-connected GNNs.

\subsection{Generic Formulation: MP-GNNs\texttt{-LSPE}}
\label{sec:generic_formulation}

{\bf Notation.} Let $\mathcal{G}= (\mathcal{V}, \mathcal{E})$ be a graph with $\mathcal{V}$ being the set of nodes and $\mathcal{E}$ the set of edges. The graph has $n=|\mathcal{V}|$ nodes and $E=|\mathcal{E}|$ edges. The connectivity of the graph is represented by the adjacency matrix $A \in \mathbb{R}^{n \times n}$ where $A_{ij} = 1$ if there exists an edge between the nodes $i$ and $j$; otherwise $A_{ij} = 0$. The degree matrix is denoted $D\in \mathbb{R}^{n \times n}$. The node features and positional features for node $i$ is denoted by $h_i$ and $p_i$ respectively, while the features for an edge between nodes $i$ and $j$ is indicated by $e_{ij}$. A GNN model is composed of three main components; an embedding layer for the input features, a stack of convolutional layers, and a final task-based layer, as in Figure \ref{fig:mpgnns_lspe_architecture}.
The layers are indexed by $\ell$ and $\ell=0$ denotes the input layer.

{\bf Standard MP-GNNs.} Considering a graph which has available node and edge features, and
these are transformed at each layer, the update equations for a conventional MP-GNN layer are defined as:
\begin{eqnarray}
\text{MP-GNNs}: \ \ \  h_i^{\ell+1} &=& \ f_{h} \Big( h_i^{\ell}, \big\{h_j^{\ell}\big\}_{j \in \mathcal{N}_i}, e_{ij}^{\ell} \Big), \ h_i^{\ell+1},h_i^{\ell}\in\mathbb{R}^{d}, \label{eqn:mpgnn_h_update}\\
    e_{ij}^{\ell+1} &=& \ f_{e} \Big( h_i^{\ell}, h_j^{\ell}, e_{ij}^{\ell}\Big), \ e_{ij}^{\ell+1},e_{ij}^{\ell}\in\mathbb{R}^{d}, \label{eqn:mpgnn_e_update}
\end{eqnarray}
where $f_h$ and $f_e$ are functions with learnable parameters, and $\mathcal{N}_i$ is the neighborhood of the node $i$. The design of functions $f_h$ and $f_e$ depends on the GNN architecture used, see \cite{zhou2020graph} for a review.
As Transformer neural networks \citep{vaswani2017attention} are a special case of MP-GNNs \citep{joshi2020transformers}, Eq. (\ref{eqn:mpgnn_h_update}) can be simplified to encompass the original Transformers by dropping the edge features and making the graph fully connected.

{\bf Input features and initialization.} The node and edge features at layer $\ell=0$ are produced by a linear embedding of available input node and edge features denoted respectively by $h_i^\textrm{in} \in \mathbb{R}^{d_v}, e_{ij}^\textrm{in} \in \mathbb{R}^{d_e}$: $h_i^{\ell=0} = \textrm{LL}_h (h_i^\textrm{in}) = A^0 h_i^\textrm{in} + a^0 \in\mathbb{R}^{d}, e_{ij}^{\ell=0} = \textrm{LL}_e (e_{ij}^\textrm{in}) = B^0 e_{ij}^\textrm{in} + b^0 \in\mathbb{R}^{d}$, 
where $A^0 \in \mathbb{R}^{d \times d_v}$, $B^0 \in \mathbb{R}^{d \times d_e}$ and $a^0,b^0 \in \mathbb{R}^d$ are the learnable parameters of the linear layers.

{\bf Positional Encoding.} Existing MP-GNNs that integrate positional information usually 
propose to concatenate the PE with the input node features, similarly to Transformers \citep{vaswani2017attention}:
\begin{align}
    \text{MP-GNNs\texttt{-PE}}: \ \ \  h_i^{\ell+1} =& \ f_{h} \Big( h_i^{\ell}, \big\{h_j^{\ell}\big\}_{j \in \mathcal{N}_i}, e_{ij}^{\ell} \Big), \ h_i^{\ell+1},h_i^{\ell}\in\mathbb{R}^{d}, \label{eqn:mpgnn-pe}\\
    e_{ij}^{\ell+1} =& \ f_{e} \Big( h_i^{\ell}, h_j^{\ell}, e_{ij}^{\ell}\Big), \ e_{ij}^{\ell+1},e_{ij}^{\ell}\in\mathbb{R}^{d}, \label{eqn:mpgnn_e_update_PE}\\
    \textrm{with initial } h_i^{\ell=0} =& \textrm{ LL}_h \left(\left[\!\!\begin{array}{c} h_i^\textrm{in} \\ p_i^\textrm{in} \\ \end{array}\!\!\right]\right) = D^0 \left[\!\!\begin{array}{c} h_i^\textrm{in} \\ p_i^\textrm{in} \\ \end{array}\!\!\right] + d^0 \in\mathbb{R}^{d}, \label{eqn:mpgnn-pe-input}\\
    \textrm{and } e_{ij}^{\ell=0} =& \textrm{ LL}_e (e_{ij}^\textrm{in}) = B^0 e_{ij}^\textrm{in} + b^0 \in\mathbb{R}^{d},
\end{align}
where $p_i^\textrm{in}\in \mathbb{R}^k$ is the input PE of node $i$, $D^0 \in \mathbb{R}^{d \times (d_v+k)}, d^0 \in \mathbb{R}^d$ are parameters for the linear transformation. Such architecture merges the positional and structural representations together. It has the advantage to keep the same linear complexity for learning, but it does not allow the positional representation to be changed and better adjusted to the task at hand.

{\bf Decoupling position and structure in MP-GNNs.} We decouple the positional information from the structural information such that both representations are learned separately resulting in an architecture with \textbf{L}earnable \textbf{S}tructural and \textbf{P}ositional \textbf{E}ncodings, which we call \textbf{MP-GNNs\texttt{-LSPE}}. The layer update equations are defined as:
\vspace{-12pt}
\begin{align}
    \text{MP-GNNs}\texttt{-LSPE}: \ \ \  h_i^{\ell+1} =& \ f_{h} \left(
    \left[\!\!\begin{array}{c} h_i^{\ell} \\ p_i^{\ell} \\ \end{array}\!\!\right],
    \left\{\!
    \left[\!\!\begin{array}{c} h_j^{\ell} \\ p_j^{\ell} \\ \end{array}\!\!\right]
\!\right\}_{j \in \mathcal{N}_i}, e_{ij}^{\ell} \right), \ h_i^{\ell+1},h_i^{\ell}\in\mathbb{R}^{d},\label{eqn:mpgnn_lspe_h_update}\\
    e_{ij}^{\ell+1} =& \ f_{e} \Big( h_i^{\ell}, h_j^{\ell}, e_{ij}^{\ell}\Big), \ e_{ij}^{\ell+1},e_{ij}^{\ell}\in\mathbb{R}^{d},\label{eqn:mpgnn_lspe_e_update}\\
    p_i^{\ell+1} =& \ f_p \Big( p_i^{\ell}, \big\{ p_j^{\ell} \big\}_{j \in \mathcal{N}_i}, e_{ij}^{\ell}\Big),\ p_i^{\ell+1},p_i^{\ell}\in\mathbb{R}^{d},\label{eqn:mpgnn_lspe_p_update}
\end{align}
The difference of this architecture with the standard MP-GNNs is the addition of the positional representation update Eq. (\ref{eqn:mpgnn_lspe_p_update}), along with the concatenation of these learnable PEs with the node structural features, Eq. (\ref{eqn:mpgnn_lspe_h_update}). As we will see in the next section, the design of the message-passing function $f_p$ follows the same analytical form of $f_h$ but with the use of the $\tanh$ activation function to allow positive and negative values for the positional coordinates. It should be noted that the inclusion of the edge features, $e_{ij}^{\ell}$ in the $h$ or $p$ update is optional as several MP-GNNs do not include edge features in their $h$ updates. Nevertheless, the architecture we present is made as generic so as to be used for future extensions in a convenient way.

{\bf Definition of initial PE.}
The choice of the initial PE
is critical. In this work, we consider two PEs: Laplacian PE (LapPE) and Random Walk PE (RWPE). 
LapPE are defined in Section \ref{sec:related_work_graphpe} as $p^\textrm{LapPE}_i = \left[ \begin{array}{c} U_{i1}, U_{i2}, \cdots, U_{ik} \end{array} \right]\in\mathbb{R}^{k}.$
LapPE provide a unique node representation and are distance-sensitive w.r.t. the Euclidean norm. However, they are limited by the sign ambiguity, which requires random sign flipping during training for the network to learn this invariance \citep{dwivedi2020benchmarking}. 

\begin{figure}[!t]
\centering
  \begin{subfigure}{0.36\linewidth}
  \centering
    \includegraphics[width=\linewidth]{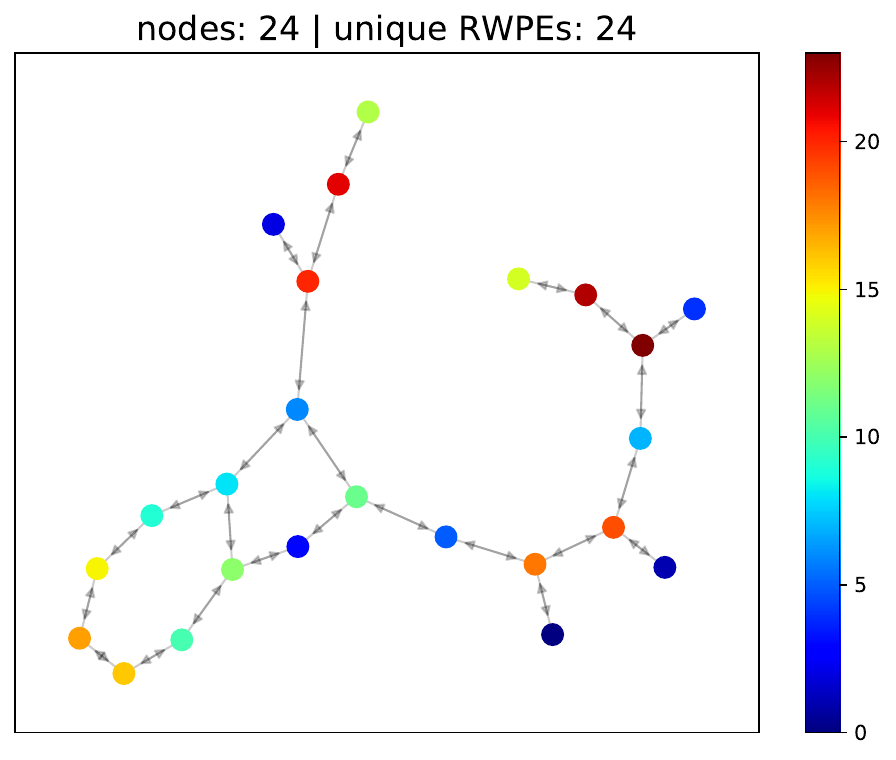}
    \vspace{-14pt}
    \caption{ZINC molecule (val index 91)}
    \label{fig:graph_id_91}
  \end{subfigure}
  \begin{subfigure}{0.36\linewidth}
  \centering
    \includegraphics[width=\linewidth]{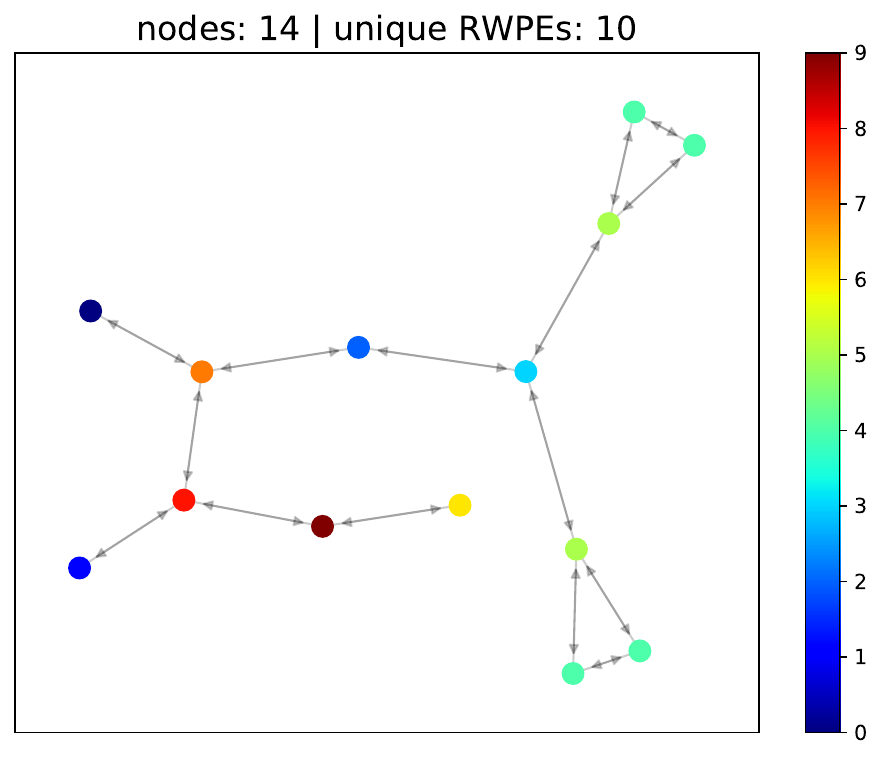}
    \vspace{-14pt}
    \caption{ZINC molecule (val index 212)}
    \label{fig:graph_id_212}
  \end{subfigure}
  \vspace{-5pt}
  \caption{Sample graph plots from the ZINC validation set with each node color in a graph representing a unique RWPE vector, when $k=24$. The corresponding graph ids, the number of nodes in the graphs and the number of unique RWPEs are labelled against the figures.}
  \label{fig:uniqueness_RWPE_graphs_main}
\end{figure}

Inspired by \cite{li2020distance}, we propose RWPE, a PE based on the random walk (RW) diffusion process (although other graph diffusions can be considered s.a. PageRank \citep{mialon2021graphit}). Formally, RWPE are defined with $k$-steps of random walk as:
\begin{eqnarray}
p^\textrm{RWPE}_i &=& \left[ \begin{array}{c} \textrm{RW}_{ii}, \textrm{RW}_{ii}^2, \cdots, \textrm{RW}_{ii}^k \end{array} \right]\in\mathbb{R}^{k}, \label{eqn:init_pe_2} 
\end{eqnarray}
where $\textrm{RW} = AD^{-1}$ is the random walk operator. In contrast of \cite{li2020distance} which uses the full matrix $\text{RW}_{ij}$ for all pairwise nodes, we adopt a low-complexity usage of the random walk matrix by considering only the landing probability of a node $i$ to itself, i.e. $\text{RW}_{ii}$. Note that these PE do not suffer from the sign ambiguity of LapPE, so the network is not required to learn additional invariance.
RWPE provide a unique node representation under the condition that each node has a unique $k$-hop topological neighborhood for a sufficient large $k$. This assumption can be discussed. If we consider synthetic strongly regular graphs like the CSL graphs \citep{murphy2019relational}, then all nodes in a graph have the same RWPE for any $k$ value, since they are isomorphic by construction. However, despite RWPE being the same for all nodes in a graph, these PE are unique for each class of isomorphic graphs,  resulting in a perfect classification of the CSL dataset, see Section \ref{sec:supplementary_power}.
For graphs such as Decalin and Bicyclopentyl \citep{sato2020survey}, nodes which are not isomorphic receive different RWPE for $k\geq 5$, also in Section \ref{sec:supplementary_power}. Finally, for real-world graphs like ZINC molecules, most nodes receive a unique node representation for $k\geq 24$, see Figure \ref{fig:uniqueness_RWPE_graphs_main} for an illustration, where the two molecules have $100\%$ and $71.43\%$ unique RWPEs respectively. Section \ref{sec:supplementary_unique_pe} presents a detailed study.

Experimentally, we will show that RWPE outperform LapPE, suggesting that learning the sign invariance is more difficult (as there exist $2^k$ possible sign flips for each graph) than not exactly having unique node representation for each node. As mentioned above for CSL, RWPE are related to the problem of graph isomorphism and higher-order node interactions. Precisely, iterating the random walk operator for a suitable number of steps allows coloring non-isomorphic nodes, thus distinguishing several cases of non-isomorphic graphs on which the 1-WL test, and equivalently MP-GNNs, fail s.a. the CSL, Decalin and Bicyclopentyl graphs. We refer to Section \ref{sec:supplementary_algo} for a formal presentation of the iterative algorithm. Finally, the initial PE of the network is obtained by embedding the LapPE or RWPE into a $d$-dimensional feature vector:
\begin{eqnarray}
    p_i^{\ell=0} &=& \textrm{LL}_p (p_i^\textrm{PE}) = C^0 p_i^\textrm{PE} + c^0 \in\mathbb{R}^{d}, \ \ \ \ \ \ \text{where} \ C^0 \in \mathbb{R}^{d \times k}, c^0 \in \mathbb{R}^d.
    \label{eqn:init_pe}
\end{eqnarray}

{\bf Positional loss.} As we separate the learning of the structual and positional representations, it is possible to consider a specific positional encoding loss along with the task loss. A natural candidate is the Laplacian eigenvector loss \citep{belkin2003laplacian, lai2014splitting} that enforces the PE to form a coordinate system constrained by the graph topology. As such, the final loss function of \textbf{MP-GNNs\texttt{-LSPE}} is composed of two terms:
\begin{align}
    \text{Loss} =& \ \text{Loss}_{\text{Task}}\left(\left[\!\!\begin{array}{c} h^{\ell=L} \\ p^{\ell=L} \\ \end{array}\!\!\right]\right)  + \alpha \  \text{Loss}_{\text{LapEig}}(p^{\ell=L}),\label{eqn:final_loss}
\end{align}
where $h^{\ell=L}\in \mathbb{R}^{n \times d}, p^{\ell=L}\in \mathbb{R}^{n \times k}$, $k$ is the dimension of learned PE, $\ell=L$ is the final GNN layer, and $\alpha>0$ an hyper-parameter. 
Observe also that we enforce the final positional vectors $p^{\ell=L}$ to have centered and unit norm $\textrm{ with }  \textrm{mean}(p^{\ell=L}_{\cdot,k})=0, \ \|p^{\ell=L}_{\cdot,k}\|=1, \ \forall k$ to better approximate the Laplacian eigenvector loss defined by $\text{Loss}_{\text{LapEig}}(p) = \ \frac{1}{k} \ \text{trace}\big(p^T \Delta p\big) + \frac{\lambda}{k} \ \big\|p^Tp - \textrm{I}_k \big\|_{F}^2$ with $\lambda>0$ and $\|\cdot\|^2_F$ being the Frobenius norm.


\subsection{Instances of \texttt{LSPE} with MP-GNNs and Transformer GNNs}
\label{sec:instances_lspe}

We instantiate two classes of GNN architectures, both \textit{sparse} MP-GNNs and \textit{fully-connected} Transformer GNNs using our proposed \texttt{LSPE} framework. For sparse MP-GNNs, we consider GatedGCN \citep{bresson2017residual} and PNA \citep{corso2020principal}, while we extend the recently developed SAN \citep{kreuzer2021rethinking} and GraphiT \citep{mialon2021graphit} with \texttt{LSPE} to develop Transformer-\texttt{LSPE} architectures.
We briefly demonstrate here how a GNN can be instantiated using \texttt{LSPE} (Eqs. (\ref{eqn:mpgnn_lspe_h_update}-\ref{eqn:mpgnn_lspe_p_update})) by developing GatedGCN\texttt{-LSPE} (Eqs. (\ref{eqn:gatedcn_lspe_1_main}-\ref{eqn:gatedgcn_lspe_3_main})), while the complete equations for the four models are defined in Section \ref{sec:model_eqns_supplementary} of the supplementary material, given the space constraint.

{\bf GatedGCN\texttt{-LSPE}:}
Originally, GatedGCNs are sparse MP-GNNs equipped with a soft-attention mechanism that is able to learn adaptive edge gates to improve the message aggregation step of GCN networks \citep{kipf2017semi}. Our proposed extension of this model with \texttt{LSPE} is defined as:
\begin{align}
    h^{\ell+1}, e^{\ell+1}, p^{\ell+1} &= \text{GatedGCN}\texttt{-LSPE}\Big(h^{\ell}, e^{\ell}, p^{\ell}\Big), \ h\in\mathbb{R}^{n\times d}, e\in\mathbb{R}^{E\times d}, p\in\mathbb{R}^{n\times d},\\
    \text{with} \ \ h_i^{\ell+1} &= h_i^{\ell} + \text{ReLU}\Big( \text{BN} \Big( A_1^{\ell} \left[ \!\!\begin{array}{c} h_i^\ell \\ p_i^\ell \\ \end{array} \!\!\right] + \sum_{j \in \mathcal{N}(i)} \eta_{ij}^{\ell} \odot A_2^{\ell} \left[ \!\!\begin{array}{c} h_j^\ell \\ p_j^\ell \\ \end{array} \!\!\right] \Big) \Big), \label{eqn:gatedcn_lspe_1_main}\\
e_{ij}^{\ell+1} &= e_{ij}^{\ell} + \text{ReLU}\big(\text{BN}\big(\hat{\eta}_{ij}^{\ell}\big)\big),\label{eqn:gatedgcn_lspe_2_main}\\
p_i^{\ell+1} &= p_i^{\ell} + \tanh \Big( C_1^{\ell}p_i^{\ell} + \sum_{j \in \mathcal{N}(i)} \eta_{ij}^{\ell} \odot C_2^{\ell}
  p_j^{\ell} \Big), \label{eqn:gatedgcn_lspe_3_main} 
\end{align}
where $\eta_{ij}^{\ell} = \sigma\big(\hat{\eta}_{ij}^{\ell}\big)/\big( \sum_{j' \in \mathcal{N}(i)} \sigma\big(\hat{\eta}_{ij'}^{\ell}\big) + \epsilon\big),  
\hat{\eta}_{ij}^{\ell} = B_1^{\ell}h_i^{\ell} + B_2^{\ell}h_j^{\ell} + B_3^{\ell}e_{ij}^{\ell}$, 
$h_i^{\ell}, e_{ij}^{\ell}, p_i^{\ell}, {\eta}_{ij}^{\ell}, \hat{\eta}_{ij}^{\ell} \in \mathbb{R}^{d}, A_1^{\ell}, A_2^{\ell} \in \mathbb{R}^{d \times 2d}$ and $B_1^{\ell}, B_2^{\ell}, B_3^{\ell}, C_1^{\ell}, C_2^{\ell} \in \mathbb{R}^{d \times d}$. Notice the $p$-update in Eq. (\ref{eqn:gatedgcn_lspe_3_main}) follows the same analytical form as the $h$-update in Eq. (\ref{eqn:gatedcn_lspe_1_main}) except for the difference in activation function, and omission of BN, which was not needed in our experiments.

\section{Numerical Experiments}
\label{sec:numerical_experiments}
We evaluate the proposed MPGNNs\texttt{-LSPE} architecture on the instances of sparse GNNs and Transformer GNNs defined in Section \ref{sec:instances_lspe} (all models 
are
presented in Section \ref{sec:model_eqns_supplementary}), using PyTorch \citep{paszke2019pytorch} and DGL \citep{wang2019deep} on standard molecular benchmarks -- ZINC \citep{irwin2012zinc}, OGBG-MOLTOX21 and OGBG-MOLPCBA \citep{hu2020ogb}. 
ZINC and MOLTOX21 are of medium scale with 12K and 7.8K graphs respectively, whereas MOLPCBA is of large scale with 437.9K graphs.
These datasets, each having a global graph-level property to be predicted, consist of molecules which are represented as graphs of atoms as nodes and bonds between the atoms as edges.
Additionally, we evaluate our architecture on three non-molecular graph datasets to show the usefulness of \texttt{LSPE} on any graph domain in general, see Section \ref{sec:experiments_non_molecular} in the supplementary.

\subsection{Datasets and Experimental Settings}
\label{sec:experiments_datasets_and_settings}
\textbf{ZINC} is a graph regression dataset where the property to be predicted for a graph is its constrained solubility which is a vital chemical property in molecular design \citep{jin2018junction}. We use the 12,000 subset of the dataset with the same splitting defined in \cite{dwivedi2020benchmarking}. Mean Absolute Error (MAE) of the property being regressed is the evaluation metric. 
\textbf{OGBG-MOLTOX21} is a multi-task binary graph classification dataset where a qualitative (active/inactive) binary label is predicted against 12 different toxicity measurements for each molecular graph \citep{tox, wu2018moleculenet}. We use the scaffold-split version of the dataset included in OGB \citep{hu2020ogb} that consists of 7,831 graphs. ROC-AUC averaged across the tasks is the evaluation metric.
\textbf{OGBG-MOLPCBA} is also a multi-task binary graph classification dataset from OGB where an active/inactive binary label is predicted for 128 bioassays \citep{wang2012pubchem, wu2018moleculenet}. It has 437,929 graphs with scaffold-split and the evaluation metric is Average Precision (AP) averaged over the tasks.

To evaluate different instantiations of our proposed MPGNNs\texttt{-LSPE}, we follow the same benchmarking protocol in \cite{dwivedi2020benchmarking} to fairly compare several models on a fixed number of 500k model parameters, for ZINC. We relax the model sizes to larger parameters for evaluation on the two OGB datasets as observed being practised on their leaderboards \citep{hu2020ogb}. The total size of parameters of each model, including the number of layers used, are indicated in the respective 
experiment
tables, with the remaining implementation details 
included in supplementary Section \ref{sec:supplementary_implementation}.

\subsection{Results and Discussion}
\label{sec:experiments_results}

The results of all our experiments on different instances of \texttt{LSPE} along with performance without using PE are presented in Table \ref{tab:all_lspe_expts} whereas the comparison of the best results from Table \ref{tab:all_lspe_expts} with baseline models and SOTA is shown in Table \ref{tab:comparison_sota}. We now summarize our observations and insights.

\begin{table}[!htb]
    \centering
    \caption{Results on the ZINC, OGBG-MOLTOX21 and OGBG-MOLPCBA datasets. All scores are averaged over 4 runs with 4 different seeds. \textbf{Bold}: GNN's best score, \textbf{\best{Red}}: Dataset's best score.}
    \scalebox{0.724}{
    \begin{tabular}{lrrcc|cccccc}
        \toprule
        \parbox[t]{2mm}{\multirow{14}{*}{\rotatebox[origin=c]{90}{ZINC}}} & \textbf{Model} &  \textbf{Init PE} & \textbf{\texttt{LSPE}} & \textbf{PosLoss} & \textbf{$L$} & \textbf{\#Param} & \textbf{TestMAE$\pm$s.d.} & \textbf{TrainMAE$\pm$s.d.} & \textbf{Epochs} & \textbf{Epoch/Total} \\
        \midrule
        & GatedGCN & \textbf{x} & \textbf{x} &\textbf{x} & 16 & 504309 & 0.251$\pm$0.009 & 0.025$\pm$0.005 & 440.25 & 8.76s/1.08hr\\
        & GatedGCN & \textbf{LapPE} & \textbf{x} & \textbf{x} & 16 & 505011 & 0.202$\pm$0.006 & 0.033$\pm$0.003 & 426.00 & 8.91s/1.22hr\\
        & GatedGCN & \textbf{RWPE} & \textbf{\checkmark} & \textbf{x} & 16 & 522870 & 0.093$\pm$0.003 & 0.014$\pm$0.003 & 440.75 & 15.17s/1.99hr\\
        & GatedGCN & \textbf{RWPE} & \textbf{\checkmark} & \textbf{\checkmark} & 16 & 522870 & \textbf{\best{0.090$\pm$0.001}} & 0.013$\pm$0.004 & 460.50 & 33.06s/4.39hr\\
        \cmidrule{2-11}
        & PNA & \textbf{x} & \textbf{x} & \textbf{x} & 16 & 369235 & 0.141$\pm$0.004 & 0.020$\pm$0.003 & 451.25 & 79.67s/10.03hr\\
        & PNA & \textbf{RWPE} & \textbf{\checkmark} & \textbf{x} & 16 & 503061 & \textbf{0.095$\pm$0.002} & 0.022$\pm$0.002 & 462.25 & 127.69s/16.61hr\\
        \cmidrule{2-11}
        & SAN & \textbf{x} & \textbf{x} & \textbf{x} & 10 & 501314 & 0.181$\pm$0.004 & 0.017$\pm$0.004 & 433.50 & 74.33s/9.23hr\\
        & SAN & \textbf{RWPE} & \textbf{\checkmark} & \textbf{x} & 10 & 588066 & \textbf{0.104$\pm$0.004} & 0.016$\pm$0.002 & 462.50 & 134.74s/17.23hr\\
        \cmidrule{2-11}
        & GraphiT & \textbf{x} & \textbf{x} & \textbf{x} & 10 & 501313 & 0.181$\pm$0.006 & 0.021$\pm$0.003 & 493.25 & 63.54s/9.37hr \\
        & GraphiT & \textbf{RWPE} & \textbf{\checkmark} & \textbf{x} & 10 & 588065 & \textbf{0.106$\pm$0.002} & 0.028$\pm$0.002 & 420.50 & 125.39s/14.84hr\\
        \toprule
        \parbox[t]{2mm}{\multirow{14}{*}{\rotatebox[origin=c]{90}{MOLTOX21}}} & \textbf{Model} &  \textbf{Init PE} & \textbf{\texttt{LSPE}} & \textbf{PosLoss} & \textbf{$L$} & \textbf{\#Param} & \textbf{TestAUC$\pm$s.d.} & \textbf{TrainAUC$\pm$s.d.} & \textbf{Epochs} & \textbf{Epoch/Total} \\
        \midrule
        & GatedGCN & \textbf{x} & \textbf{x} & \textbf{x} & 8 & 1003739 & 0.772$\pm$0.006 & 0.933$\pm$0.010 & 304.25 & 5.12s/0.46hr\\
        & GatedGCN & \textbf{LapPE} & \textbf{x} & \textbf{x} & 8 & 1004355 & 0.774$\pm$0.007 & 0.921$\pm$0.006 & 275.50 & 5.23s/0.48hr\\
        & GatedGCN & \textbf{RWPE} & \textbf{\checkmark} & \textbf{x} & 8 & 1063821 & \best{\textbf{0.775$\pm$0.003}} & 0.906$\pm$0.003 & 246.50 & 5.99s/0.63hr\\
        \cmidrule{2-11}
        & PNA & \textbf{x} & \textbf{x} & \textbf{x} & 8 & 5244849 & 0.755$\pm$0.008 & 0.876$\pm$0.014 & 214.75 & 6.25s/0.38hr \\
        & PNA & \textbf{RWPE} & \textbf{\checkmark} & \textbf{x} & 8 & 5357393 & \textbf{0.761$\pm$0.007} & 0.871$\pm$0.009 & 215.50 & 7.61s/0.56hr\\
        & PNA & \textbf{RWPE} & \textbf{\checkmark} & \textbf{\checkmark} & 8 & 5357393 & \textbf{0.758$\pm$0.003} & 0.875$\pm$0.012 & 194.25 & 18.09s/1.07hr\\
        \cmidrule{2-11}
        & SAN & \textbf{x} & \textbf{x} & \textbf{x} & 10 & 957871 & \textbf{0.744$\pm$0.007} & 0.915$\pm$0.015 & 279.75 & 18.06s/1.44hr \\
        & SAN & \textbf{RWPE} & \textbf{\checkmark} & \textbf{x} & 10 & 1051017 & 0.744$\pm$0.008 & 0.918$\pm$0.018 & 281.75 & 30.82s/2.84hr\\
        \cmidrule{2-11}
        & GraphiT & \textbf{x} & \textbf{x} & \textbf{x} & 10 & 957870 & 0.743$\pm$0.003 & 0.919$\pm$0.023 & 276.50 & 16.73s/1.36hr \\
        & GraphiT & \textbf{RWPE} & \textbf{\checkmark} & \textbf{x} & 10 & 1051788 & \textbf{0.746$\pm$0.010} & 0.934$\pm$0.016 & 279.75 & 27.92s/2.57hr\\
        \toprule
        \parbox[t]{2mm}{\multirow{8}{*}{\rotatebox[origin=c]{90}{MOLPCBA}}} & \textbf{Model} &  \textbf{Init PE} & \textbf{\texttt{LSPE}} & \textbf{PosLoss} & \textbf{$L$} & \textbf{\#Param} & \textbf{TestAP$\pm$s.d.} & \textbf{TrainAP$\pm$s.d.} & \textbf{Epochs} & \textbf{Epoch/Total} \\
        \midrule
        & GatedGCN & \textbf{x} & \textbf{x} & \textbf{x} & 8 & 1008263 & 0.262$\pm$0.001 & 0.401$\pm$0.057 & 190.50 & 149.10s/7.91hr\\
        & GatedGCN & \textbf{LapPE} & \textbf{x} & \textbf{x} & 8 & 1008879 & 0.266$\pm$0.002 & 0.391$\pm$0.003 & 177.00 & 152.94s/8.29hr\\
        & GatedGCN & \textbf{RWPE} & \textbf{\checkmark} & \textbf{x} & 8 & 1068721 & \textbf{0.267$\pm$0.002} & 0.403$\pm$0.006 & 181.00 & 206.43s/11.64hr \\
        \cmidrule{2-11}
        & PNA & \textbf{x} & \textbf{x} & \textbf{x} & 4 & 6550839 & 0.279$\pm$0.003 & 0.448$\pm$0.004 & 129.25 & 174.75s/6.34hr \\
        & PNA & \textbf{RWPE} & \textbf{\checkmark} & \textbf{x} & 4 & 6521029 & \best{\textbf{0.284$\pm$0.002}} & 0.383$\pm$0.005 & 320.00 & 201.05s/22.99hr\\
        \bottomrule
    \end{tabular}
    }
    \label{tab:all_lspe_expts}
\end{table}

{\bf No PE results in lowest performance.} In Table \ref{tab:all_lspe_expts}, the GNNs which do not use PE tend to give the worse performance on all the three datasets. This finding is aligned to the recent literature (Sec. \ref{sec:related_work_graphpe}) that has guided research towards powerful PE methods for expressive GNNs. Besides, it can be observed that the extent of poor performance of models without PE against using a PE (LapPE or \texttt{LSPE}) is greater for ZINC than the two OGBG-MOL* datasets used. This difference can be explained by the fact that ZINC
features are purely atom and bond descriptors whereas OGB-MOL* features consist additional information that is informative of e.g. 
if an atom is in ring, among others.

{\bf \texttt{LSPE} boosts the capabilities of existing GNNs.}
Both sparse GNNs and Transformer GNNs are improved significantly when they are augmented with \texttt{LSPE} having RWPE as initial PE, see Table \ref{tab:all_lspe_expts}. For instance, the best GNN without PE for ZINC, i.e. PNA, gives an improvement of $32.62\%$ ($0.095$ vs. $0.141$) when \texttt{LSPE} is used to learn the structural and positional representations in a decoupled manner. On other GNNs, this boost is even higher, see GatedGCN\texttt{-LSPE} which shows a gain of
$64.14\%$ ($0.090$ vs. $0.251$).
On MOLTOX21, PNA\texttt{-LSPE} improves $0.79\%$ ($0.761$ vs. $0.755$) over PNA while the remaining models show either minor gains or attain the same performance when not using PE. This consistent trend is also observed for MOLPCBA where \texttt{LSPE} boosts PNA by $1.79\%$.

{\bf Sparse vs. Transformer GNNs.} When we compare the performance of sparse GNNs (GatedGCN, PNA) against Transformer GNNs (SAN, GraphiT) augmented with \texttt{LSPE} in Table \ref{tab:all_lspe_expts}, the performance of the sparse GNNs is surprisingly better than the latter, despite Transformer GNNs being theoretically well-posed to counter the limitations of long-range interactions of the former. Notably, the evaluation of our proposed architecture, in this work, is on molecular graphs on which the information among local structures seems to be the most critical, diminishes the need of full attention. This also aligns with the insight put forward in \cite{kreuzer2021rethinking} where the SAN, a Transformer model, benefited less from full attention on molecules. Beyond molecular graphs, there may be other domains where Transformer GNNs could give better performance, but still these would not scale in view of the quadratic computational complexity. Indeed, it is important to notice the much lesser training times of sparse GNNs compared to Transformer GNNs in Table \ref{tab:all_lspe_expts}.

{\bf \texttt{LSPE} improves the state-of-the-art for domain-agnostic GNNs.} When we compare the best performing instantiation of the \texttt{LSPE} from Table \ref{tab:all_lspe_expts} with baseline GNN models from the literature on the three benchmark datasets, our proposed architecture improves the SOTA on 
ZINC,
while achieving SOTA-comparable performance on 
remaining datasets,
see Table \ref{tab:comparison_sota}. On ZINC, GatedGCN\texttt{-LSPE} surpasses most baselines by a large margin to give a test MAE of $0.090$
which is an improvement of 
$35.25\%$ and $26.23\%$
respectively over the two recent-most Transformer based GNNs, SAN and Graphormer. On MOLTOX21, GatedGCN\texttt{-LSPE} reports a test ROC-AUC score of $0.7754$ which is similar to the best baseline GIN ($0.7757$) that uses virtual node (VN). Finally, \texttt{LSPE} enables PNA to achieve comparable performance to SOTA on MOLPCBA while boosting its performance when no PE was used. We note here that ZINC scores can even be boosted beyond \texttt{LSPE}'s SOTA when domain expertise is used \citep{bouritsas2020improving, bodnar2021weisfeiler} while Graphormer \citep{ying2021transformers} achieved the top score on MOLPCBA when pre-trained on a very large (3.8M graphs) 
dataset. To ensure fair comparison with other scores, we did not use these two results in Table \ref{tab:comparison_sota}.

\begin{table}[!t]
\centering
\caption{Comparison of our best \texttt{LSPE} results from Table \ref{tab:all_lspe_expts} with baselines and state-of-the-art GNNs (Sec. \ref{sec:models_comparison_sota}) on each dataset. For ZINC, all the scores in Table \ref{tab:comparison_zinc} are the models with the $\sim$500k parameters. The scores on OGBG-MOL* in Tables \ref{tab:comparison_moltox21} and \ref{tab:comparison_molpcba} are taken from the OGB project and its leaderboards \citep{hu2020ogb}, where models have different number of parameters.}
    \vspace{-6pt}
    \scalebox{0.75}{
    \begin{subtable}{.43\linewidth}
      \centering
      \caption{ZINC}
        \begin{tabular}{rc}
        \toprule
        \textbf{Model} & \textbf{Test MAE}\\
        \midrule
        GCN & 0.367$\pm$0.011\\
        GAT & 0.384$\pm$0.007\\
        GatedGCN-LapPE & 0.202$\pm$0.006\\
        GT & 0.226$\pm$0.014\\
        SAN & 0.139$\pm$0.006\\
        Graphormer & 0.122$\pm$0.006\\
        \midrule
        GatedGCN\texttt{-LSPE} & \textbf{0.090$\pm$0.001}\\
        \bottomrule
        \end{tabular}
        \label{tab:comparison_zinc}
    \end{subtable}%
    \begin{subtable}{.43\linewidth}
      \centering
      \caption{OGBG-MOLTOX21}
        \begin{tabular}{rc}
        \toprule
        \textbf{Model} & \textbf{Test ROC-AUC}\\
        \midrule
        GCN & 0.7529$\pm$0.0069\\
        GCN-VN & 0.7746$\pm$0.0086\\
        GIN & 0.7491$\pm$0.0051\\
        GIN-VN & \textbf{0.7757$\pm$0.0062}\\
        GatedGCN-LapPE & 0.7743$\pm$0.0073\\
        \midrule
        GatedGCN\texttt{-LSPE} & \textbf{0.7754$\pm$0.0032}\\
        \bottomrule
        \end{tabular}
        \label{tab:comparison_moltox21}
    \end{subtable}
    \begin{subtable}{.43\linewidth}
      \centering
      \caption{OGBG-MOLPCBA}
        \begin{tabular}{rc}
        \toprule
        \textbf{Model} & \textbf{Test AP}\\
        \midrule
        GIN & 0.2266$\pm$0.0028\\
        GIN-VN & 0.2703$\pm$0.0023\\
        DeeperGCN-VN & 0.2781$\pm$0.0038\\
        PNA & 0.2838$\pm$0.0035\\
        DGN & 0.2885$\pm$0.0030\\
        PHC-GNN & \textbf{0.2947$\pm$0.0026}\\
        \midrule
        PNA\texttt{-LSPE} & 0.2840$\pm$0.0021\\
        \bottomrule
        \end{tabular}
        \label{tab:comparison_molpcba}
    \end{subtable}
    }
    \label{tab:comparison_sota}
    \vspace{-16pt}
\end{table}

{\bf On Positional loss.}
It can be observed in Table \ref{tab:all_lspe_expts} that the positional loss 
Eq. (\ref{eqn:final_loss}),
further pushes the best \texttt{LSPE} score on ZINC slightly from $0.093$ to $0.090$, while on MOLTOX21 it only improves the train score though obtaining comparable test performance.
We will investigate a more consistent positional loss in a future work.

Finally, we would like to highlight the generic nature of our proposed architecture which can be applied to any MP-GNN in practice as demonstrated by four diverse GNNs in this work.

\subsection{Ablation Studies}
\label{sec:experiments_ablation_studies}
Through ablation studies, we show -- i) the usefulness of learning positional representation at every layer vs. simply injecting a pre-computed positional encoding in the input layer,
and ii) the selection of the number of $k$ for the steps in RWPE in the proposed \texttt{LSPE} architecture.

{\bf Learning PE at every layer provides the best performance.} In Table \ref{tab:ablation_LSPE_PE}, GatedGCN-RWPE corresponds to the model where LapPE are replaced with $k$-dim pre-computed random walk features at the first layer, and the PE are not updated in the subsequent layers. First, we observe a significant leap in performance (from $0.202$ to $0.122$) when the RWPE are injected in place of LapPE at the first layer, suggesting that RWPE could encode better positional information in GNNs as they are not limited by the sign ambiguity of LapPE. See Section \ref{sec:supplementary_power} in the supplementary material for an example of RWPE's representation power. Note that the injection of RWPE at the final layer instead of the input layer gives poor performance. The reason behind the better performance of concatenating RWPE at the input layer is to inform the GNN aggregation function of the node positions in order to distinguish them in the case of graph symmetries like isomorphic nodes.

Now, if we observe the training performance, GatedGCN-RWPE leads to an overfit on ZINC. However, when the positional representations are also updated, the overfit is considerably alleviated improving the test score to $0.100$. Finally, when we further fuse the learned positional features at the final layer with the structural features, Eq. (\ref{eqn:final_loss}), the model achieves the best  MAE test of $0.093$. This study justifies how the GNN model learns best when the positional representations can be tuned and better adjusted to the learning task being dealt with.

\begin{table}[!t]
    \centering
        \caption{Comparing the final \texttt{LSPE} architecture against simpler models which add pre-computed PE at input layer (or final layer) of a GNN, using GatedGCN model on ZINC. The column `Final $h$' denotes whether only the node structural features are used as final node features (denoted by $h^L$), or are concatenated with (i) node positional features (denoted by $[h^L, p^L]$) at the final layer, (ii) pre-computed RWPE (denoted by $[h^L, \textbf{RWPE}]$).}
    \scalebox{0.68}{
    \begin{tabular}{rrcc|cccccc}
        \toprule
        \textbf{Model} &  \textbf{Init PE} & \textbf{\texttt{LSPE}} & \textbf{Final $h$}  & \textbf{$L$} & \textbf{\#Param} & \textbf{Test MAE$\pm$s.d.} & \textbf{Train MAE$\pm$s.d.} & \textbf{\#Epochs} & \textbf{Epoch/Total} \\
        \midrule
        GatedGCN & \textbf{x} & \textbf{x} & $h^L$ & 16 & 504309 & 0.251$\pm$0.009 & 0.025$\pm$0.005 & 440.25 & 8.76s/1.08hr\\
        GatedGCN & \textbf{LapPE} & \textbf{x} & $h^L$ & 16 & 505011 & 0.202$\pm$0.006 & 0.033$\pm$0.003 & 426.00 & 8.91s/1.22hr\\
        \midrule
        GatedGCN & \textbf{RWPE} & \textbf{x} & $h^L$ & 16 & 505947 & 0.122$\pm$0.003 & 0.013$\pm$0.003 & 436.25 & 9.14s/1.28hr\\
        GatedGCN & \textbf{x} & \textbf{x} & $[h^L, \textbf{RWPE}]$ & 16 & 515249 & 0.249$\pm$0.012 & 0.024$\pm$0.002 & 437.50 & 10.05s/1.55hr\\
        \midrule
        GatedGCN & \textbf{LapPE} & \textbf{\checkmark} & $h^L$ & 16 & 516722 & 0.202$\pm$0.008 & 0.032$\pm$0.005 & 405.25 & 15.10s/1.84hr\\
        GatedGCN & \textbf{LapPE} & \textbf{\checkmark} & $[h^L, p^L]$ & 16 & 520734 & 0.196$\pm$0.008 & 0.023$\pm$0.004 & 454.00 & 15.22s/2.06hr\\
        \midrule
        GatedGCN & \textbf{RWPE} & \textbf{\checkmark} & $h^L$ & 16 & 518150 & 0.100$\pm$0.006 & 0.018$\pm$0.012 & 395.00 & 15.09s/1.73hr\\
        GatedGCN & \textbf{RWPE} & \textbf{\checkmark} & $[h^L, p^L]$ & 16 & 522870 & 0.093$\pm$0.003 & 0.014$\pm$0.003 & 440.75 & 15.17s/1.99hr\\
        \bottomrule
\end{tabular}
    }
    \label{tab:ablation_LSPE_PE}
    \vspace{-10pt}
\end{table}

{\bf The choice of $k$ steps to initialize RWPE.}
\label{sec:experiments_ablation_k}
In Figure \ref{fig:k_plots} (see Section \ref{sec:figure_k_study}), we study the effect of choosing a suitable number of $k$ steps for the random walk features that are used as initial positional encoding in Section \ref{sec:generic_formulation}.
This value $k$ is also used to set the final dimension of the learned positional representation in the last layer. Numerical experiments show the best values of $k$ to be $20$ and $16$ for ZINC with GatedGCN-LSPE and OGBG-MOLTOX21 with PNA-LSPE respectively, which are larger values from what was used in \cite{li2020distance} ($k=3,4$) where the RW features are treated as distance encoding. The difference of $k$ value is due to two reasons. First, the proposed RWPE requires to use a large $k$ value to possibly provide a unique node representation with different $k$-hop neighborhoods. Second, \cite{li2020distance} not only uses $\text{RW}^k_{ii}$ but also considers all pairwise $\text{RW}^k_{ij}$ between nodes $i$ and $j$ in a target set of nodes, which increases the computational complexity.

\section{Conclusion}
\label{sec:conclusion}
This work presents a novel approach to learn structural and positional representations separately in a graph neural network. The resultant architecture, \texttt{LSPE} enables a principled and effective learning of these two key properties that make GNN representation even more expressive. Main design components of \texttt{LSPE} are -- i) higher-order position informative random walk features as PE initialization, ii) decoupling positional representations at every GNN layer, and iii) the fusion of the structural and positional features finally to generate hybrid features for the learning task. We observe a consistent increase of performance across several instances of our model on the benchmark datasets used for evaluation. Our architecture is simple and universal to be used with any sparse GNNs or Transformer GNNs as demonstrated by two sparse GNNs and two fully connected Transformer based GNNs in our numerical experiments. Given the importance of incorporating expressive positional encodings to theoretically improve GNNs as seen in the recent literature, we believe this paper provides a useful architectural framework that can be considered when developing future models which improve graph positional encodings, for both GNNs and Transformers.


\subsubsection*{Ethics Statement}
In this work, we present an approach to improve neural network methods for graphs by considering efficient learnable positional encoding while keeping the linear complexity of the model w.r.t to the number of nodes. This improves the cost of training such models, as contrast to some previous works that improved GNNs at the cost of higher-order tensor computation. We discover another insight that the linear complexity models (sparse GNNs) can outperform quadratic complexity models (Transformers). Consequently, one beneficial impact of our work is that its use can reduce GPU and computational resources, eventually contributing to minimizing the adverse effect of deep learning training on environment. However, the method we propose belongs to a class of architectures that can be used on malicious applications since the internet and several of the processes in its ecosystem can be represented in form of graphs. To prevent such applications, ethical guidelines can be set and enforced which constraint the usage of our proposed model.


\subsubsection*{Reproducibility Statement}

The authors support and advocate the principles of open science and reproducible research. The algorithms and architectures proposed in this work are open-sourced in a free and public code repository with easy-to-use scripts to reproduce different experiments and evaluations presented. The tables included in the paper mention critical details on the number of layers and the total number of model parameters that are trained. Similarly, the visualization and illustrations presented in the main paper as well as the supplementary material contain the exact details on the dataset examples (such as index) used. Finally, a detailed table consisting of several hyperparameters used for the experiments are included in the supplementary, ensuring the reproducibility of the results discussed in this work. 

\subsubsection*{Acknowledgments}
XB is supported by NRF Fellowship NRFF2017-10 and NUS-R-252-000-B97-133. This research is supported by Nanyang Technological University, under SUG Grant (020724-00001). VPD would like to thank Andreea Deac for her helpful feedback, Quan Gan for his support on the DGL library, Gabriele Corso for answering questions related to the PNA model, and Chaitanya K. Joshi for useful comments. Finally, the authors would like to thank the anonymous reviewers for their helpful suggestions and feedbacks.

\bibliography{iclr2021_conference}

\begin{thebibliography}{70}
\providecommand{\natexlab}[1]{#1}
\providecommand{\url}[1]{\texttt{#1}}
\expandafter\ifx\csname urlstyle\endcsname\relax
  \providecommand{\doi}[1]{doi: #1}\else
  \providecommand{\doi}{doi: \begingroup \urlstyle{rm}\Url}\fi

\bibitem[Alon \& Yahav(2020)Alon and Yahav]{alon2020bottleneck}
Uri Alon and Eran Yahav.
\newblock On the bottleneck of graph neural networks and its practical
  implications.
\newblock \emph{arXiv preprint arXiv:2006.05205}, 2020.

\bibitem[Azizian \& Lelarge(2020)Azizian and Lelarge]{azizian2020expressive}
Wa{\"\i}ss Azizian and Marc Lelarge.
\newblock Expressive power of invariant and equivariant graph neural networks.
\newblock \emph{arXiv preprint arXiv:2006.15646}, 2020.

\bibitem[Bapst et~al.(2020)Bapst, Keck, Grabska-Barwi{\'n}ska, Donner, Cubuk,
  Schoenholz, Obika, Nelson, Back, Hassabis, et~al.]{bapst2020unveiling}
Victor Bapst, Thomas Keck, A~Grabska-Barwi{\'n}ska, Craig Donner, Ekin~Dogus
  Cubuk, Samuel~S Schoenholz, Annette Obika, Alexander~WR Nelson, Trevor Back,
  Demis Hassabis, et~al.
\newblock Unveiling the predictive power of static structure in glassy systems.
\newblock \emph{Nature Physics}, 16\penalty0 (4):\penalty0 448--454, 2020.

\bibitem[Beani et~al.(2021)Beani, Passaro, L{\'e}tourneau, Hamilton, Corso, and
  Li{\`o}]{beani2021directional}
Dominique Beani, Saro Passaro, Vincent L{\'e}tourneau, Will Hamilton, Gabriele
  Corso, and Pietro Li{\`o}.
\newblock Directional graph networks.
\newblock In \emph{International Conference on Machine Learning}, pp.\
  748--758. PMLR, 2021.

\bibitem[Belkin \& Niyogi(2003)Belkin and Niyogi]{belkin2003laplacian}
Mikhail Belkin and Partha Niyogi.
\newblock Laplacian eigenmaps for dimensionality reduction and data
  representation.
\newblock \emph{Neural computation}, 15\penalty0 (6):\penalty0 1373--1396,
  2003.

\bibitem[Bengio et~al.(2021)Bengio, Lodi, and Prouvost]{bengio2021machine}
Yoshua Bengio, Andrea Lodi, and Antoine Prouvost.
\newblock Machine learning for combinatorial optimization: a methodological
  tour d’horizon.
\newblock \emph{European Journal of Operational Research}, 290\penalty0
  (2):\penalty0 405--421, 2021.

\bibitem[Bodnar et~al.(2021)Bodnar, Frasca, Otter, Wang, Li{\`o}, Mont{\'u}far,
  and Bronstein]{bodnar2021weisfeiler}
Cristian Bodnar, Fabrizio Frasca, Nina Otter, Yu~Guang Wang, Pietro Li{\`o},
  Guido Mont{\'u}far, and Michael Bronstein.
\newblock Weisfeiler and lehman go cellular: Cw networks.
\newblock \emph{arXiv preprint arXiv:2106.12575}, 2021.

\bibitem[Bouritsas et~al.(2020)Bouritsas, Frasca, Zafeiriou, and
  Bronstein]{bouritsas2020improving}
Giorgos Bouritsas, Fabrizio Frasca, Stefanos Zafeiriou, and Michael~M
  Bronstein.
\newblock Improving graph neural network expressivity via subgraph isomorphism
  counting.
\newblock \emph{arXiv preprint arXiv:2006.09252}, 2020.

\bibitem[Bresson \& Laurent(2017)Bresson and Laurent]{bresson2017residual}
Xavier Bresson and Thomas Laurent.
\newblock Residual gated graph convnets.
\newblock \emph{arXiv preprint arXiv:1711.07553}, 2017.

\bibitem[Cappart et~al.(2021)Cappart, Ch{\'e}telat, Khalil, Lodi, Morris, and
  Veli{\v{c}}kovi{\'c}]{cappart2021combinatorial}
Quentin Cappart, Didier Ch{\'e}telat, Elias Khalil, Andrea Lodi, Christopher
  Morris, and Petar Veli{\v{c}}kovi{\'c}.
\newblock Combinatorial optimization and reasoning with graph neural networks.
\newblock \emph{arXiv:2102.09544}, 2021.

\bibitem[Chen et~al.(2019)Chen, Chen, Villar, and Bruna]{chen2019equivalence}
Zhengdao Chen, Lei Chen, Soledad Villar, and Joan Bruna.
\newblock On the equivalence between graph isomorphism testing and function
  approximation with gnns.
\newblock \emph{Advances in neural information processing systems}, 2019.

\bibitem[Corso et~al.(2020)Corso, Cavalleri, Beaini, Li{\`o}, and
  Veli{\v{c}}kovi{\'c}]{corso2020principal}
Gabriele Corso, Luca Cavalleri, Dominique Beaini, Pietro Li{\`o}, and Petar
  Veli{\v{c}}kovi{\'c}.
\newblock Principal neighbourhood aggregation for graph nets.
\newblock \emph{Advances in Neural Information Processing Systems}, 33, 2020.

\bibitem[Cranmer et~al.(2019)Cranmer, Xu, Battaglia, and
  Ho]{cranmer2019learning}
Miles~D Cranmer, Rui Xu, Peter Battaglia, and Shirley Ho.
\newblock Learning symbolic physics with graph networks.
\newblock \emph{arXiv preprint arXiv:1909.05862}, 2019.

\bibitem[Cui et~al.(2021)Cui, Lu, Li, and Yang]{cui2021positional}
Hejie Cui, Zijie Lu, Pan Li, and Carl Yang.
\newblock On positional and structural node features for graph neural networks
  on non-attributed graphs.
\newblock \emph{arXiv preprint arXiv:2107.01495}, 2021.

\bibitem[Defferrard et~al.(2016)Defferrard, Bresson, and
  Vandergheynst]{defferrard2016convolutional}
Micha{\"e}l Defferrard, Xavier Bresson, and Pierre Vandergheynst.
\newblock Convolutional neural networks on graphs with fast localized spectral
  filtering.
\newblock In \emph{NIPS}, 2016.

\bibitem[Derrow-Pinion et~al.(2021)Derrow-Pinion, She, Wong, Lange, Hester,
  Perez, Nunkesser, Lee, Guo, Battaglia, Gupta, Li, Xu, Sanchez-Gonzalez, Li,
  and Veli\v{c}kovi\'{c}]{derrowpinion2021traffic}
Austin Derrow-Pinion, Jennifer She, David Wong, Oliver Lange, Todd Hester, Luis
  Perez, Marc Nunkesser, Seongjae Lee, Xueying Guo, Peter~W Battaglia, Vishal
  Gupta, Ang Li, Zhongwen Xu, Alvaro Sanchez-Gonzalez, Yujia Li, and Petar
  Veli\v{c}kovi\'{c}.
\newblock {Traffic Prediction with Graph Neural Networks in Google Maps}.
\newblock 2021.

\bibitem[Dufter et~al.(2021)Dufter, Schmitt, and
  Sch{\"u}tze]{dufter2021position}
Philipp Dufter, Martin Schmitt, and Hinrich Sch{\"u}tze.
\newblock Position information in transformers: An overview.
\newblock \emph{arXiv preprint arXiv:2102.11090}, 2021.

\bibitem[Dwivedi \& Bresson(2021)Dwivedi and
  Bresson]{dwivedi2021generalization}
Vijay~Prakash Dwivedi and Xavier Bresson.
\newblock A generalization of transformer networks to graphs.
\newblock In \emph{AAAI Workshop on Deep Learning on Graphs: Methods and
  Applications}, 2021.

\bibitem[Dwivedi et~al.(2020)Dwivedi, Joshi, Laurent, Bengio, and
  Bresson]{dwivedi2020benchmarking}
Vijay~Prakash Dwivedi, Chaitanya~K Joshi, Thomas Laurent, Yoshua Bengio, and
  Xavier Bresson.
\newblock Benchmarking graph neural networks.
\newblock \emph{arXiv preprint arXiv:2003.00982}, 2020.

\bibitem[Fowlkes et~al.(2004)Fowlkes, Belongie, Chung, and
  Malik]{fowlkes2004spectral}
Charless Fowlkes, Serge Belongie, Fan Chung, and Jitendra Malik.
\newblock Spectral grouping using the nystrom method.
\newblock \emph{IEEE transactions on pattern analysis and machine
  intelligence}, 26\penalty0 (2):\penalty0 214--225, 2004.

\bibitem[Gaudelet et~al.(2020)Gaudelet, Day, Jamasb, Soman, Regep, Liu, Hayter,
  Vickers, Roberts, Tang, et~al.]{gaudelet2020utilising}
Thomas Gaudelet, Ben Day, Arian~R Jamasb, Jyothish Soman, Cristian Regep,
  Gertrude Liu, Jeremy~BR Hayter, Richard Vickers, Charles Roberts, Jian Tang,
  et~al.
\newblock Utilising graph machine learning within drug discovery and
  development.
\newblock \emph{arXiv preprint arXiv:2012.05716}, 2020.

\bibitem[Gilmer et~al.(2017)Gilmer, Schoenholz, Riley, Vinyals, and
  Dahl]{gilmer2017neural}
Justin Gilmer, Samuel~S Schoenholz, Patrick~F Riley, Oriol Vinyals, and
  George~E Dahl.
\newblock Neural message passing for quantum chemistry.
\newblock In \emph{International Conference on Machine Learning}, pp.\
  1263--1272. PMLR, 2017.

\bibitem[Hamilton et~al.(2017)Hamilton, Ying, and
  Leskovec]{hamilton2017inductive}
William~L Hamilton, Rex Ying, and Jure Leskovec.
\newblock Inductive representation learning on large graphs.
\newblock In \emph{Proceedings of the 31st International Conference on Neural
  Information Processing Systems}, pp.\  1025--1035, 2017.

\bibitem[Hochreiter \& Schmidhuber(1997)Hochreiter and
  Schmidhuber]{hochreiter1997long}
Sepp Hochreiter and J{\"u}rgen Schmidhuber.
\newblock Long short-term memory.
\newblock \emph{Neural computation}, 9\penalty0 (8):\penalty0 1735--1780, 1997.

\bibitem[Hu et~al.(2020)Hu, Fey, Zitnik, Dong, Ren, Liu, Catasta, and
  Leskovec]{hu2020ogb}
Weihua Hu, Matthias Fey, Marinka Zitnik, Yuxiao Dong, Hongyu Ren, Bowen Liu,
  Michele Catasta, and Jure Leskovec.
\newblock Open graph benchmark: Datasets for machine learning on graphs.
\newblock \emph{arXiv preprint arXiv:2005.00687}, 2020.

\bibitem[Ioffe \& Szegedy(2015)Ioffe and Szegedy]{ioffe2015batch}
Sergey Ioffe and Christian Szegedy.
\newblock Batch normalization: Accelerating deep network training by reducing
  internal covariate shift.
\newblock In \emph{International conference on machine learning}, pp.\
  448--456. PMLR, 2015.

\bibitem[Irwin et~al.(2012)Irwin, Sterling, Mysinger, Bolstad, and
  Coleman]{irwin2012zinc}
John~J Irwin, Teague Sterling, Michael~M Mysinger, Erin~S Bolstad, and Ryan~G
  Coleman.
\newblock Zinc: a free tool to discover chemistry for biology.
\newblock \emph{Journal of chemical information and modeling}, 52\penalty0
  (7):\penalty0 1757--1768, 2012.

\bibitem[Islam et~al.(2020)Islam, Jia, and Bruce]{Islam2020How}
Md~Amirul Islam, Sen Jia, and Neil D.~B. Bruce.
\newblock How much position information do convolutional neural networks
  encode?
\newblock In \emph{International Conference on Learning Representations}, 2020.

\bibitem[Jin et~al.(2018)Jin, Barzilay, and Jaakkola]{jin2018junction}
Wengong Jin, Regina Barzilay, and Tommi Jaakkola.
\newblock Junction tree variational autoencoder for molecular graph generation.
\newblock In \emph{International conference on machine learning}, pp.\
  2323--2332. PMLR, 2018.

\bibitem[Joshi(2020)]{joshi2020transformers}
Chaitanya Joshi.
\newblock Transformers are graph neural networks.
\newblock \emph{The Gradient}, 2020.

\bibitem[Khasahmadi et~al.(2020)Khasahmadi, Hassani, Moradi, Lee, and
  Morris]{Khasahmadi2020Memory-Based}
Amir~Hosein Khasahmadi, Kaveh Hassani, Parsa Moradi, Leo Lee, and Quaid Morris.
\newblock Memory-based graph networks.
\newblock In \emph{International Conference on Learning Representations}, 2020.
\newblock URL \url{https://openreview.net/forum?id=r1laNeBYPB}.

\bibitem[Kipf \& Welling(2017)Kipf and Welling]{kipf2017semi}
Thomas~N. Kipf and Max Welling.
\newblock Semi-supervised classification with graph convolutional networks.
\newblock In \emph{International Conference on Learning Representations
  (ICLR)}, 2017.

\bibitem[Kreuzer et~al.(2021)Kreuzer, Beaini, Hamilton, L{\'e}tourneau, and
  Tossou]{kreuzer2021rethinking}
Devin Kreuzer, Dominique Beaini, William~L Hamilton, Vincent L{\'e}tourneau,
  and Prudencio Tossou.
\newblock Rethinking graph transformers with spectral attention.
\newblock \emph{arXiv preprint arXiv:2106.03893}, 2021.

\bibitem[Krizhevsky et~al.(2009)Krizhevsky, Hinton,
  et~al.]{krizhevsky2009learning}
Alex Krizhevsky, Geoffrey Hinton, et~al.
\newblock Learning multiple layers of features from tiny images.
\newblock 2009.

\bibitem[Lai \& Osher(2014)Lai and Osher]{lai2014splitting}
Rongjie Lai and Stanley Osher.
\newblock A splitting method for orthogonality constrained problems.
\newblock \emph{Journal of Scientific Computing}, 58\penalty0 (2):\penalty0
  431--449, 2014.

\bibitem[Le et~al.(2021)Le, Bertolini, No{\'e}, and
  Clevert]{le2021parameterized}
Tuan Le, Marco Bertolini, Frank No{\'e}, and Djork-Arn{\'e} Clevert.
\newblock Parameterized hypercomplex graph neural networks for graph
  classification.
\newblock \emph{arXiv preprint arXiv:2103.16584}, 2021.

\bibitem[LeCun et~al.(1998)LeCun, Bottou, Bengio, and
  Haffner]{lecun1998gradient}
Yann LeCun, L{\'e}on Bottou, Yoshua Bengio, and Patrick Haffner.
\newblock Gradient-based learning applied to document recognition.
\newblock \emph{Proceedings of the IEEE}, 86\penalty0 (11):\penalty0
  2278--2324, 1998.

\bibitem[Li et~al.(2019)Li, Muller, Thabet, and Ghanem]{li2019deepgcns}
Guohao Li, Matthias Muller, Ali Thabet, and Bernard Ghanem.
\newblock Deepgcns: Can gcns go as deep as cnns?
\newblock In \emph{Proceedings of the IEEE/CVF International Conference on
  Computer Vision}, pp.\  9267--9276, 2019.

\bibitem[Li et~al.(2020{\natexlab{a}})Li, Xiong, Thabet, and
  Ghanem]{li2020deepergcn}
Guohao Li, Chenxin Xiong, Ali Thabet, and Bernard Ghanem.
\newblock Deepergcn: All you need to train deeper gcns.
\newblock \emph{arXiv preprint arXiv:2006.07739}, 2020{\natexlab{a}}.

\bibitem[Li et~al.(2020{\natexlab{b}})Li, Wang, Wang, and
  Leskovec]{li2020distance}
Pan Li, Yanbang Wang, Hongwei Wang, and Jure Leskovec.
\newblock Distance encoding: Design provably more powerful neural networks for
  graph representation learning.
\newblock \emph{Advances in Neural Information Processing Systems}, 33,
  2020{\natexlab{b}}.

\bibitem[Li et~al.(2020{\natexlab{c}})Li, Qian, Zhang, and Liu]{li2020graph}
Yang Li, Buyue Qian, Xianli Zhang, and Hui Liu.
\newblock Graph neural network-based diagnosis prediction.
\newblock \emph{Big Data}, 8\penalty0 (5):\penalty0 379--390,
  2020{\natexlab{c}}.

\bibitem[Loukas(2020)]{Loukas2020What}
Andreas Loukas.
\newblock What graph neural networks cannot learn: depth vs width.
\newblock In \emph{International Conference on Learning Representations}, 2020.

\bibitem[Maron et~al.(2018)Maron, Ben-Hamu, Shamir, and
  Lipman]{maron2018invariant}
Haggai Maron, Heli Ben-Hamu, Nadav Shamir, and Yaron Lipman.
\newblock Invariant and equivariant graph networks.
\newblock \emph{arXiv preprint arXiv:1812.09902}, 2018.

\bibitem[Maron et~al.(2019)Maron, Ben-Hamu, Serviansky, and
  Lipman]{maron2019provably}
Haggai Maron, Heli Ben-Hamu, Hadar Serviansky, and Yaron Lipman.
\newblock Provably powerful graph networks.
\newblock \emph{arXiv preprint arXiv:1905.11136}, 2019.

\bibitem[Mialon et~al.(2021)Mialon, Chen, Selosse, and
  Mairal]{mialon2021graphit}
Gr{\'e}goire Mialon, Dexiong Chen, Margot Selosse, and Julien Mairal.
\newblock Graphit: Encoding graph structure in transformers.
\newblock \emph{arXiv preprint arXiv:2106.05667}, 2021.

\bibitem[Monti et~al.(2017)Monti, Boscaini, Masci, Rodola, Svoboda, and
  Bronstein]{monti2017geometric}
Federico Monti, Davide Boscaini, Jonathan Masci, Emanuele Rodola, Jan Svoboda,
  and Michael~M Bronstein.
\newblock Geometric deep learning on graphs and manifolds using mixture model
  cnns.
\newblock In \emph{Proceedings of the IEEE conference on computer vision and
  pattern recognition}, pp.\  5115--5124, 2017.

\bibitem[Monti et~al.(2019)Monti, Frasca, Eynard, Mannion, and
  Bronstein]{monti2019fake}
Federico Monti, Fabrizio Frasca, Davide Eynard, Damon Mannion, and Michael~M
  Bronstein.
\newblock Fake news detection on social media using geometric deep learning.
\newblock \emph{arXiv preprint arXiv:1902.06673}, 2019.

\bibitem[Morris et~al.(2019)Morris, Ritzert, Fey, Hamilton, Lenssen, Rattan,
  and Grohe]{morris2019weisfeiler}
Christopher Morris, Martin Ritzert, Matthias Fey, William~L Hamilton, Jan~Eric
  Lenssen, Gaurav Rattan, and Martin Grohe.
\newblock Weisfeiler and leman go neural: Higher-order graph neural networks.
\newblock In \emph{Proceedings of the AAAI Conference on Artificial
  Intelligence}, volume~33, pp.\  4602--4609, 2019.

\bibitem[Morris et~al.(2020)Morris, Kriege, Bause, Kersting, Mutzel, and
  Neumann]{morris2020tudataset}
Christopher Morris, Nils~M Kriege, Franka Bause, Kristian Kersting, Petra
  Mutzel, and Marion Neumann.
\newblock Tudataset: A collection of benchmark datasets for learning with
  graphs.
\newblock \emph{arXiv preprint arXiv:2007.08663}, 2020.

\bibitem[Murphy et~al.(2019)Murphy, Srinivasan, Rao, and
  Ribeiro]{murphy2019relational}
Ryan Murphy, Balasubramaniam Srinivasan, Vinayak Rao, and Bruno Ribeiro.
\newblock Relational pooling for graph representations.
\newblock In \emph{International Conference on Machine Learning}, pp.\
  4663--4673. PMLR, 2019.

\bibitem[Naor \& Stockmeyer(1995)Naor and Stockmeyer]{naor1995can}
Moni Naor and Larry Stockmeyer.
\newblock What can be computed locally?
\newblock \emph{SIAM Journal on Computing}, 24\penalty0 (6):\penalty0
  1259--1277, 1995.

\bibitem[Pal et~al.(2020)Pal, Eksombatchai, Zhou, Zhao, Rosenberg, and
  Leskovec]{pal2020pinnersage}
Aditya Pal, Chantat Eksombatchai, Yitong Zhou, Bo~Zhao, Charles Rosenberg, and
  Jure Leskovec.
\newblock Pinnersage: Multi-modal user embedding framework for recommendations
  at pinterest.
\newblock In \emph{Proceedings of the 26th ACM SIGKDD International Conference
  on Knowledge Discovery \& Data Mining}, pp.\  2311--2320, 2020.

\bibitem[Pan et~al.(2004)Pan, Yang, Faloutsos, and Duygulu]{pan2004automatic}
Jia-Yu Pan, Hyung-Jeong Yang, Christos Faloutsos, and Pinar Duygulu.
\newblock Automatic multimedia cross-modal correlation discovery.
\newblock In \emph{Proceedings of the tenth ACM SIGKDD international conference
  on Knowledge discovery and data mining}, pp.\  653--658, 2004.

\bibitem[Paszke et~al.(2019)Paszke, Gross, Massa, Lerer, Bradbury, Chanan,
  Killeen, Lin, Gimelshein, Antiga, et~al.]{paszke2019pytorch}
Adam Paszke, Sam Gross, Francisco Massa, Adam Lerer, James Bradbury, Gregory
  Chanan, Trevor Killeen, Zeming Lin, Natalia Gimelshein, Luca Antiga, et~al.
\newblock Pytorch: An imperative style, high-performance deep learning library.
\newblock \emph{Advances in neural information processing systems},
  32:\penalty0 8026--8037, 2019.

\bibitem[Sato(2020)]{sato2020survey}
Ryoma Sato.
\newblock A survey on the expressive power of graph neural networks.
\newblock \emph{arXiv preprint arXiv:2003.04078}, 2020.

\bibitem[Sato et~al.(2019)Sato, Yamada, and Kashima]{sato2019approximation}
Ryoma Sato, Makoto Yamada, and Hisashi Kashima.
\newblock Approximation ratios of graph neural networks for combinatorial
  problems.
\newblock In \emph{Advances in Neural Information Processing Systems}, pp.\
  4081--4090, 2019.

\bibitem[Srinivasan \& Ribeiro(2019)Srinivasan and
  Ribeiro]{srinivasan2019equivalence}
Balasubramaniam Srinivasan and Bruno Ribeiro.
\newblock On the equivalence between positional node embeddings and structural
  graph representations.
\newblock In \emph{International Conference on Learning Representations}, 2019.

\bibitem[Stokes et~al.(2020)Stokes, Yang, Swanson, Jin, Cubillos-Ruiz, Donghia,
  MacNair, French, Carfrae, Bloom-Ackermann, et~al.]{stokes2020deep}
Jonathan~M Stokes, Kevin Yang, Kyle Swanson, Wengong Jin, Andres Cubillos-Ruiz,
  Nina~M Donghia, Craig~R MacNair, Shawn French, Lindsey~A Carfrae, Zohar
  Bloom-Ackermann, et~al.
\newblock A deep learning approach to antibiotic discovery.
\newblock \emph{Cell}, 180\penalty0 (4):\penalty0 688--702, 2020.

\bibitem[Sukhbaatar et~al.(2016)Sukhbaatar, Fergus,
  et~al.]{sukhbaatar2016learning}
Sainbayar Sukhbaatar, Rob Fergus, et~al.
\newblock Learning multiagent communication with backpropagation.
\newblock \emph{Advances in neural information processing systems},
  29:\penalty0 2244--2252, 2016.

\bibitem[Tox21(2014)]{tox}
Tox21.
\newblock Tox21 challenge.
\newblock 2014.
\newblock URL \url{https://tripod.nih.gov/tox21/challenge/}.

\bibitem[Vaswani et~al.(2017)Vaswani, Shazeer, Parmar, Uszkoreit, Jones, Gomez,
  Kaiser, and Polosukhin]{vaswani2017attention}
Ashish Vaswani, Noam Shazeer, Niki Parmar, Jakob Uszkoreit, Llion Jones,
  Aidan~N Gomez, Lukasz Kaiser, and Illia Polosukhin.
\newblock Attention is all you need.
\newblock In \emph{NIPS}, 2017.

\bibitem[Veli{\v{c}}kovi{\'c} et~al.(2018)Veli{\v{c}}kovi{\'c}, Cucurull,
  Casanova, Romero, Li{\`o}, and Bengio]{velivckovic2018graph}
Petar Veli{\v{c}}kovi{\'c}, Guillem Cucurull, Arantxa Casanova, Adriana Romero,
  Pietro Li{\`o}, and Yoshua Bengio.
\newblock Graph attention networks.
\newblock In \emph{International Conference on Learning Representations}, 2018.

\bibitem[Wang et~al.(2019)Wang, Yu, Zheng, Gan, Gai, Ye, Li, Zhou, Huang, Ma,
  et~al.]{wang2019deep}
Minjie Wang, Lingfan Yu, Da~Zheng, Quan Gan, Yu~Gai, Zihao Ye, Mufei Li,
  Jinjing Zhou, Qi~Huang, Chao Ma, et~al.
\newblock Deep graph library: Towards efficient and scalable deep learning on
  graphs.
\newblock 2019.

\bibitem[Wang et~al.(2012)Wang, Xiao, Suzek, Zhang, Wang, Zhou, Han,
  Karapetyan, Dracheva, Shoemaker, et~al.]{wang2012pubchem}
Yanli Wang, Jewen Xiao, Tugba~O Suzek, Jian Zhang, Jiyao Wang, Zhigang Zhou,
  Lianyi Han, Karen Karapetyan, Svetlana Dracheva, Benjamin~A Shoemaker, et~al.
\newblock Pubchem's bioassay database.
\newblock \emph{Nucleic acids research}, 40\penalty0 (D1):\penalty0 D400--D412,
  2012.

\bibitem[Weisfeiler \& Leman(1968)Weisfeiler and
  Leman]{weisfeiler1968reduction}
Boris Weisfeiler and Andrei Leman.
\newblock The reduction of a graph to canonical form and the algebra which
  appears therein.
\newblock \emph{NTI Series}, 2\penalty0 (9):\penalty0 12--16, 1968.

\bibitem[Wu et~al.(2018)Wu, Ramsundar, Feinberg, Gomes, Geniesse, Pappu,
  Leswing, and Pande]{wu2018moleculenet}
Z~Wu, B~Ramsundar, EN~Feinberg, J~Gomes, C~Geniesse, AS~Pappu, K~Leswing, and
  V~Pande.
\newblock Moleculenet: a benchmark for molecular machine learning. chem sci 9:
  513--530, 2018.

\bibitem[Xu et~al.(2019)Xu, Hu, Leskovec, and Jegelka]{xu2018how}
Keyulu Xu, Weihua Hu, Jure Leskovec, and Stefanie Jegelka.
\newblock How powerful are graph neural networks?
\newblock In \emph{International Conference on Learning Representations}, 2019.

\bibitem[Ying et~al.(2021)Ying, Cai, Luo, Zheng, Ke, He, Shen, and
  Liu]{ying2021transformers}
Chengxuan Ying, Tianle Cai, Shengjie Luo, Shuxin Zheng, Guolin Ke, Di~He,
  Yanming Shen, and Tie-Yan Liu.
\newblock Do transformers really perform bad for graph representation?
\newblock \emph{arXiv preprint arXiv:2106.05234}, 2021.

\bibitem[You et~al.(2019)You, Ying, and Leskovec]{you2019position}
Jiaxuan You, Rex Ying, and Jure Leskovec.
\newblock Position-aware graph neural networks.
\newblock In \emph{International Conference on Machine Learning}, pp.\
  7134--7143. PMLR, 2019.

\bibitem[Zhou et~al.(2020)Zhou, Cui, Hu, Zhang, Yang, Liu, Wang, Li, and
  Sun]{zhou2020graph}
Jie Zhou, Ganqu Cui, Shengding Hu, Zhengyan Zhang, Cheng Yang, Zhiyuan Liu,
  Lifeng Wang, Changcheng Li, and Maosong Sun.
\newblock Graph neural networks: A review of methods and applications.
\newblock \emph{AI Open}, 1:\penalty0 57--81, 2020.

\end{thebibliography}
\bibliographystyle{iclr2021_conference}

\newpage
\appendix
\section{Supplementary}
\label{sec:supplementary}

\subsection{Distinguishing non-isomorphic graphs using Random Walk features}
\label{sec:supplementary_power}

The choice of the initial PE in our proposed architecture can be several based on graph diffusion or other related techniques. In this section, we study RWPE (Eqn. \ref{eqn:init_pe_2}) which we initialize with $k$-steps of random walk. Precisely we use a $k$-dim vector that encodes the landing probabilities of a node $i$ to itself in $1$ to $k$ steps. This initial PE vector for a node $i$ is given by $[\text{RW}_{ii}, \text{RW}^2_{ii}, \ldots, \text{RW}^k_{ii}] \in \mathbb{R}^{k}$ which is pre-computed before the model training. Here, we demonstrate that such PE vector can help distinguish i) structurally dissimilar nodes and ii) non-isomorphic graphs on which 1-WL, and equivalently MP-GNNs, fail, thus illustrating the empirically powerful nature of MPGNNs\texttt{-LSPE} that relies on this choice of positional features initialization.

\begin{figure}[h]
\includegraphics[width=0.94\linewidth]{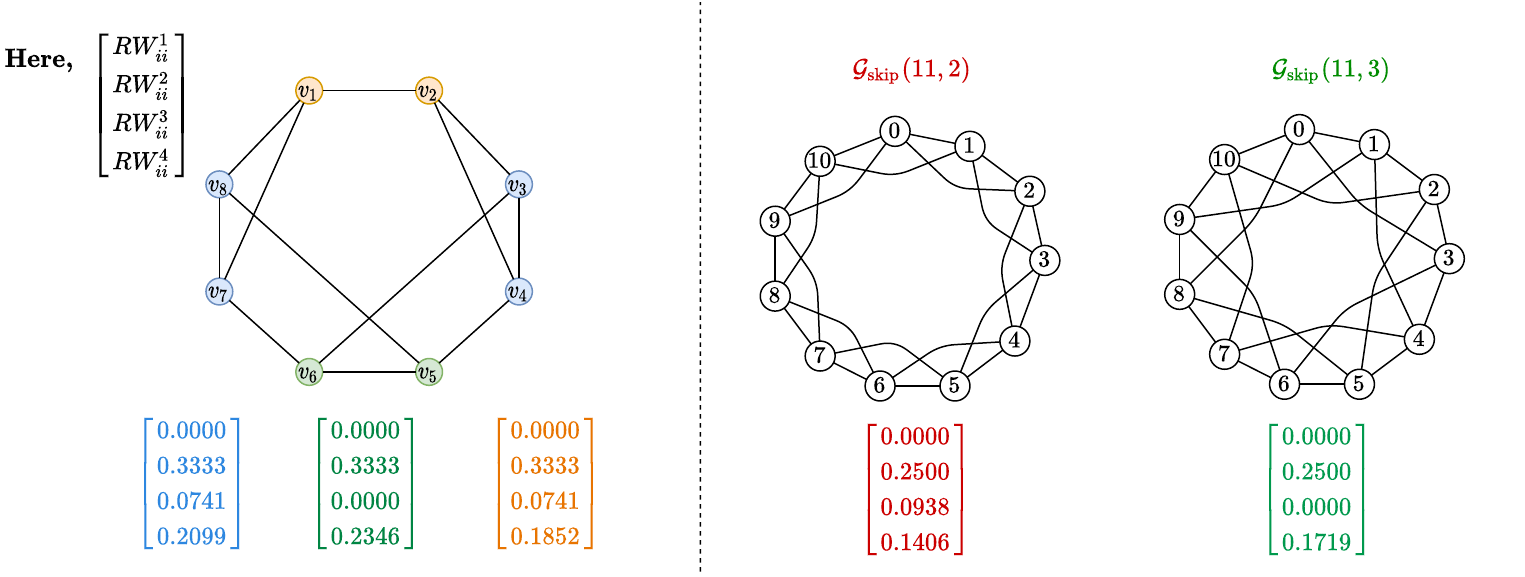}
\caption{\textbf{Left:} Example $3$-regular graph with $8$ nodes from \cite{li2020distance} where the nodes are structurally different and colored accordingly. The $4$-dim initial RWPE vector is shown against the corresponding nodes with their respective colors. \textbf{Right:} Example pair of non-isomorphic graphs with $11$ nodes and skip-links $2$ and $3$ from \cite{murphy2019relational}. Each node in a graph gets the same $4$-dim RWPE vector, and shown above in colors are the respective graphs' RWPE vectors after averaging across all the nodes.}
\label{fig:distinguish_nodes_1}
\end{figure}

\begin{figure}[h]
\centering
\includegraphics[width=0.94\linewidth]{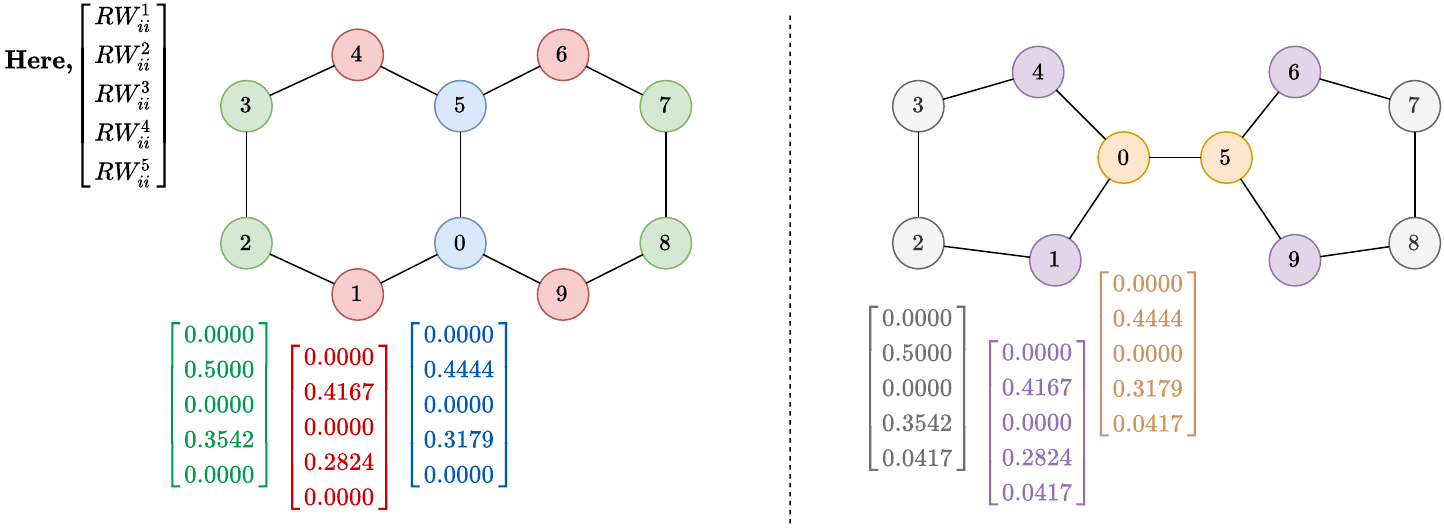}
\caption{A pair of non-isomorphic and non-regular graphs (Left: Decalin, Right: Bicyclopentyl) from \cite{sato2020survey}. The $5$-dim initial PE vector is shown against the corresponding nodes with their respective colors.}
\label{fig:distinguish_nodes_2}
\end{figure}

We show the simulation of the nodes' initial RWPE vectors on three examples in Figure \ref{fig:distinguish_nodes_1} (Left), Figure \ref{fig:distinguish_nodes_1} (Right), and Figure \ref{fig:distinguish_nodes_2} where the graphs either do not have any node attributes (Figure \ref{fig:distinguish_nodes_1}), or have the same node attributes (Figure \ref{fig:distinguish_nodes_2} where each node denotes a Carbon atom). When we apply MPGNNs on the graph in \ref{fig:distinguish_nodes_1} (Left), each node will have the same feature representation as it is a regular graph without any node attributes. However, there are structurally 3 different kinds of nodes denoted by the same number of different colors. If we initialize the PE for these nodes for $k=4$ random walk steps, we can observe that the nodes are being assigned the $4$-dim feature vectors that is consistent to their initial structural roles in the graph, thus being distinguishable. 

Similarly, Figure \ref{fig:distinguish_nodes_1} (Right) is a pair of non-isomorphic graphs from the theoretically challenging and highly symmetric Circulant Skip Link (CSL) dataset from \cite{murphy2019relational}. It can be noticed that every node in a graph here has the same structural role as the each node has edges with other nodes at same hops. However, in $\mathcal{G}_{\text{skip}}(11,2)$, the edges are between nodes at ${1,2}$ hops whereas in $\mathcal{G}_{\text{skip}}(11,3)$, the edges are between the nodes at ${1,3}$ hops, with $2$ and $3$ being the skip-links of the two graphs, respectively. In such a scenario, the node in the $\mathcal{G}_{\text{skip}}(11,2)$ gets a different $4$-dim initial PE than a node in $\mathcal{G}_{\text{skip}}(11,3)$, thus helping eventually to distinguish the two graphs when these node features are pooled to generate the graph feature vector. 

Finally, in Figure \ref{fig:distinguish_nodes_2}, a pair of non-isomorphic and non-regular graphs is shown from \cite{sato2020survey} that MPGNNs fail to distinguish. 
If we use $5$ steps of Random Walk to initialize the node's PE vector, we can observe that the two graphs can easily be distinguished. We note here that the random walk based PE initialization (RWPE) is close to one of the Distance Encoding instantiations used in \cite{li2020distance}. However, we do not require to consider pairwise scores $\text{RW}^k_{ij}$ between nodes $i$ and $j$ and any sub-set of nodes from the original graph, thus making our method less computationally demanding.

\subsection{Random Walk PE Feature And Graph Isomorphism Test}
\label{sec:supplementary_algo}

Similar to the 1-WL test for graph isomorphism \citep{weisfeiler1968reduction, morris2019weisfeiler, sato2020survey}, the RWPE can be used as a node coloring algorithm to test if two graphs are non-isomorphic, as described in Algorithm \ref{algo:rw_iso_test}. Note that this algorithm cannot guarantee that two graphs are isomorphic, like the WL test. However, our analysis in Section \ref{sec:supplementary_power} shows this algorithm to be strictly powerful than the 1-WL test as the pairs of graphs in Figure \ref{fig:distinguish_nodes_1} (Right) and in Figure \ref{fig:distinguish_nodes_2} are not distinguishable by 1-WL. Although this increase in power is being achieved without the need of maintaining colors for tuple of nodes to encode higher order interactions (as in k-WL), the algorithm's complexity is of $O(k * n^3)$ due to the matrix multiplication in Step 5 (b) and Step 5 (c), compared to $O(k * n^2)$ of 1-WL, with $k$ being the number of iterations until convergence.

\begin{algorithm}
\caption{Algorithm to decide whether a pair of graphs are not isomorphic based on random walk landing probabilities of each node to itself.}

\textbf{Input:} A pair of graphs $\mathcal{G}_1 = (\mathcal{V}_1, \mathcal{E}_1)$, $\mathcal{G}_2 = (\mathcal{V}_2, \mathcal{E}_2)$ with $n$ nodes and $e$ edges in each graph. ${A}_1 \in \mathbb{R}^{n \times n}$ and ${A}_2 \in \mathbb{R}^{n \times n}$ denote the adjacency matrices, ${D}_1 \in \mathbb{R}^{n \times n}$ and ${D}_2 \in \mathbb{R}^{n \times n}$ denote the degree matrices of graphs $\mathcal{G}_1$ and $\mathcal{G}_2$ respectively.\\
\textbf{Output:} Return ``non-isomorphic" if $\mathcal{G}_1$ and $\mathcal{G}_2$ are not isomorphic else ``possibly isomorphic".
\begin{enumerate}
    \item $M^{(0)} \ \gets \ A_1 D_1^{-1} \ \in \mathbb{R}^{n \times n}$
    \item $N^{(0)} \ \gets \ A_2 D_2^{-1} \ \in \mathbb{R}^{n \times n}$
    \item $c_u^{(0)} \ \gets \ M^{(0)}_{u,u} \quad \forall u \in \mathcal{V}_1$
    \item $d_v^{(0)} \ \gets \ N^{(0)}_{v,v} \quad \forall v \in \mathcal{V}_2$
    \item for $k = 1, 2, \cdots $ (until convergence to stationary distribution)
    \begin{enumerate}
        \item if \textsc{Hash}$\Big(\{\{c_u^{(k-1)} \in \mathbb{R}^{k} \ | \ u \in \mathcal{V}_1\}\}\Big) \neq$ \textsc{Hash}$\Big(\{\{d_v^{(k-1)} \in \mathbb{R}^{k} \ | \ v \in \mathcal{V}_2\}\}\Big)$ then return ``non-isomorphic"
        \item $M^{(k)} \ \gets \ M^{(k-1)} M^{(0)}\ \in \mathbb{R}^{n \times n}$
        \item $N^{(k)} \ \gets \ N^{(k-1)} N^{(0)}\ \in \mathbb{R}^{n \times n}$
        \item $c_u^{(k)} \ \gets \ \text{append} \ \ M^{(k)}_{u,u} \ \ \text{to} \ \ c_u^{(k-1)}\quad \forall u \in \mathcal{V}_1$
        \item $d_v^{(k)} \ \gets \ \text{append} \ \ N^{(k)}_{v,v} \ \ \text{to} \ \ d_v^{(k-1)}\quad \forall v \in \mathcal{V}_2$
    \end{enumerate}
    \item return ``possibly isomorphic"
\end{enumerate}

\label{algo:rw_iso_test}
\end{algorithm}
where \textsc{Hash} is an injective hash function and $\{\{\ldots\}\}$ denotes a multiset.

\subsection{Study of LapPE and RWPE as initial PE}
\label{sec:supplementary_unique_pe}

\begin{figure}[h]
\centering
  \begin{subfigure}{0.42\linewidth}
  \centering
    \includegraphics[width=\linewidth]{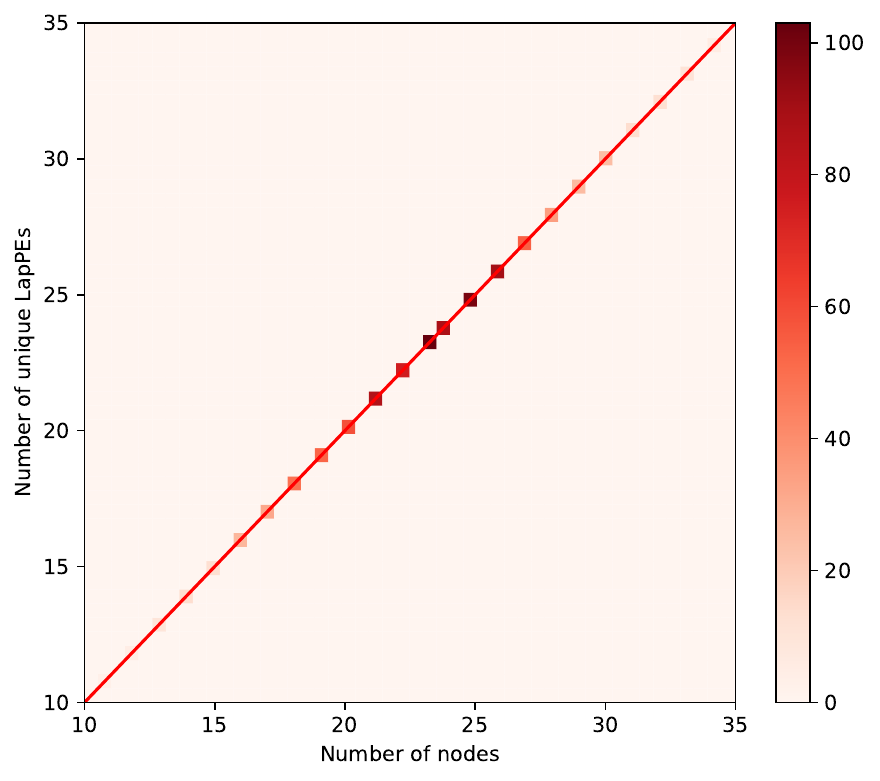}
    \vspace{-9pt}
    \caption{LapPE, $k=36$}
    \label{fig:uniqueness_LapPE}
  \end{subfigure}
  \hspace{7pt}
  \begin{subfigure}{0.42\linewidth}
  \centering
    \includegraphics[width=\linewidth]{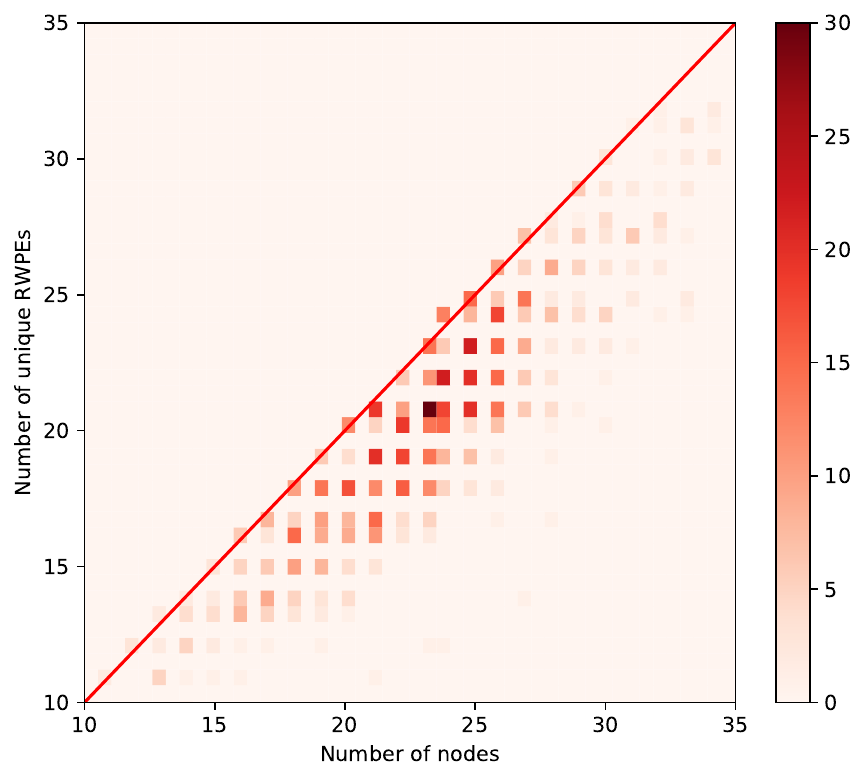}
    \vspace{-9pt}
    \caption{RWPE, $k=24$}
    \label{fig:uniqueness_RWPE}
  \end{subfigure}
  \vspace{-5pt}
  \caption{Plot of the number of nodes in a graph vs. the number of unique PE for LapPE and RWPE. A point in the plots represents a graph in the ZINC validation set (composed of $1000$ graphs) where the $x$-axis is the number of nodes, the $y$-axis is the number of unique PEs and the point intensity is the number of graphs with the same pair $(x,y)$. Besides, Fig. \ref{fig:uniqueness_LapPE} has $36$-dim LapPE (trailing dims padded with zero for a graph with $n<36$), and Fig. \ref{fig:uniqueness_RWPE} has $24$-dim RWPE.}
  \label{fig:uniqueness_PE}
\end{figure}

\begin{figure}[h]
\centering
  \begin{subfigure}{0.4\linewidth}
  \centering
    \includegraphics[width=\linewidth]{images/ZINC_valset_graph_91.pdf}
    \vspace{-12pt}
    \caption{ZINC molecule (val index 91)}
    \label{fig:graph_id_91}
  \end{subfigure}
  \hspace{6pt}
  \begin{subfigure}{0.4\linewidth}
  \centering
    \includegraphics[width=\linewidth]{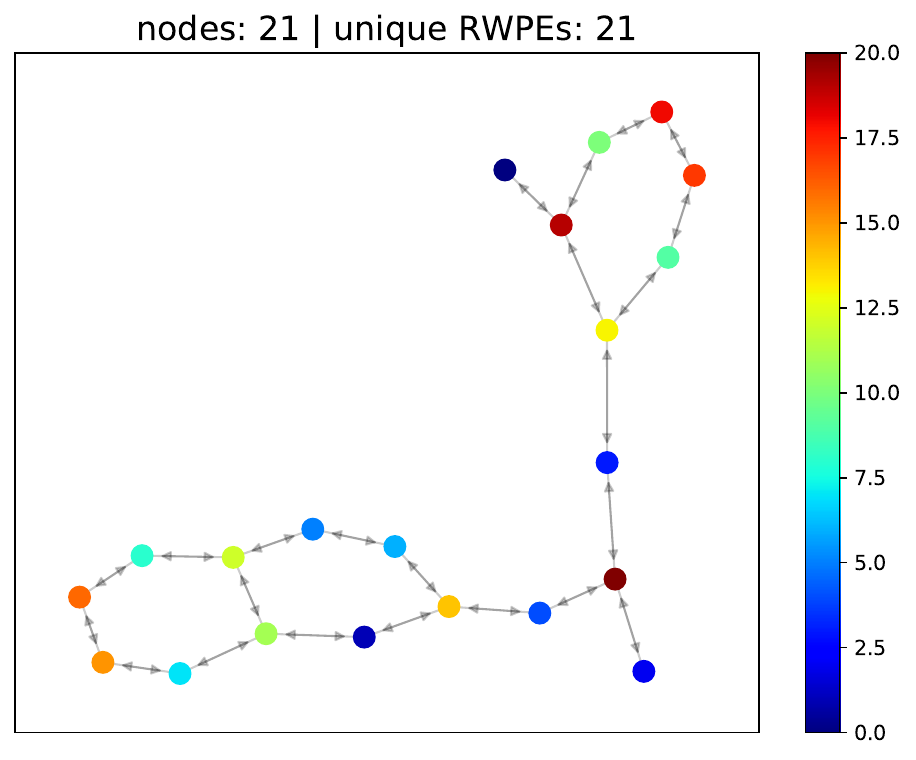}
    \vspace{-12pt}
    \caption{ZINC molecule (val index 967)}
    \label{fig:graph_id_967}
  \end{subfigure}
  \vspace{0.35cm}
  
  \begin{subfigure}{0.4\linewidth}
  \centering
    \includegraphics[width=\linewidth]{images/ZINC_valset_graph_212.pdf}
    \vspace{-12pt}
    \caption{ZINC molecule (val index 212)}
    \label{fig:graph_id_212}
  \end{subfigure}
  \hspace{6pt}
  \begin{subfigure}{0.4\linewidth}
  \centering
    \includegraphics[width=\linewidth]{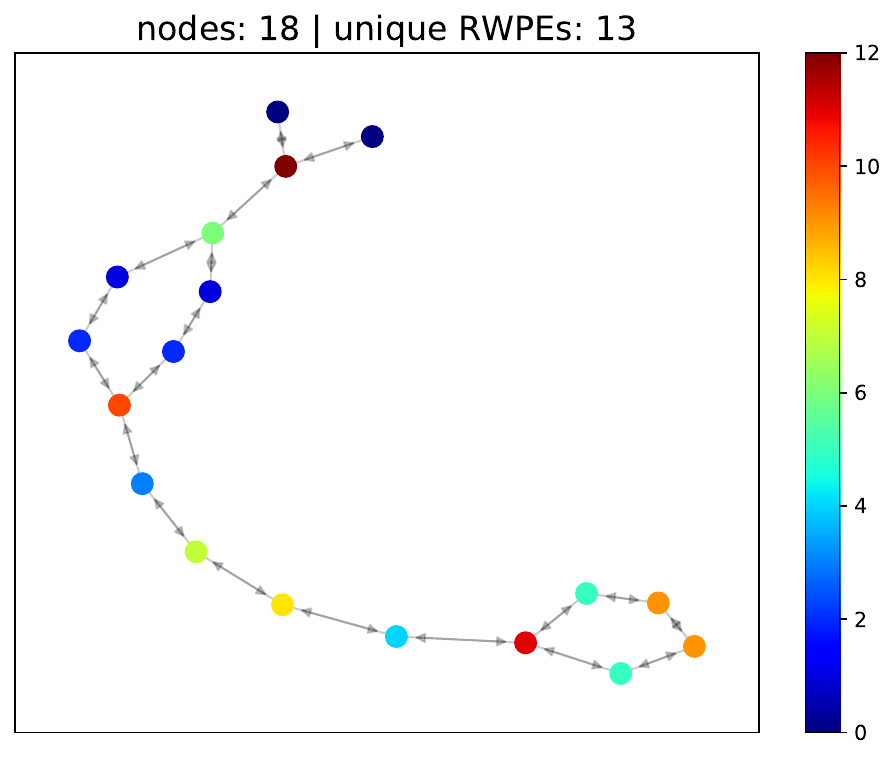}
    \vspace{-12pt}
    \caption{ZINC molecule (val index 672)}
    \label{fig:graph_id_672}
  \end{subfigure}
  \vspace{-5pt}
  \caption{Sample graph plots from the ZINC validation set with each node color in a graph representing a unique RWPE vector, when $k=24$. The corresponding graph ids, the number of nodes in the graphs and the number of unique RWPEs are labelled against the figures.}
  \label{fig:uniqueness_RWPE_graphs}
\end{figure}

Figure \ref{fig:uniqueness_PE} visualizes the uniqueness of the node representation with LapPE and RWPE (which serve as initial PE of our network) using the ZINC validation set of $1000$ real-world molecular graphs. If the initial PE is unique for each node in a graph, then the graph lies on the straight diagonal line. Figure \ref{fig:uniqueness_LapPE} shows the result for LapPE, all graphs lie on the diagonal line as Laplacian eigenvectors guarantee unique node coordinates in the Euclidean transformed space. Figure \ref{fig:uniqueness_RWPE} presents the result for RWPE. We observe that not all, but a large amount of ZINC molecular graphs stay close to the straight line, showing that most graphs have a large amount of nodes with unique RWPE. For example, there are 30 graphs with 24 nodes having 21 unique RWPE, equivalent to 87.5\% of nodes with unique PE.

Additionally, we visualize four sample graph plots from the ZINC validation set in Figure \ref{fig:uniqueness_RWPE_graphs} where the first two graphs have completely unique RWPE features, while the next two graphs have partially unique RWPEs ($71.43\%$ and $72.22\%$ respectively). The visualization assigns a unique node color for each unique  RWPE representation. Therefore, graphs in Figures \ref{fig:graph_id_91} and \ref{fig:graph_id_967} are plotted with each node assigned to a unique color based on their RWPE features, and graphs in Figures \ref{fig:graph_id_212} and \ref{fig:graph_id_672} are represented with 10 and 13 unique colors respectively corresponding to their number of unique RWPE representations. In particular, observe the green-shade colored nodes in Figure \ref{fig:graph_id_212} (top and bottom-right) as well as blue-shade (mid-left) and orange-shade (bottom-right) colored nodes in Figure \ref{fig:graph_id_672}. We can easily see that the nodes with the same color are isomorphic in the graph, i.e. their $k$-hop structural neighborhoods are the same for values $k\geq 11$.

We remind that RWPE provides a unique node representation under the condition that each node have a unique $k$-hop topological neighborhood for a sufficient large $k$. While this condition is experimentally true for most nodes, it is not always satisfied. But despite this approximation, for a sufficiently large number $k$ of random walk iterations, RWPE is still able to capture global higher-order positioning of nodes that is used as initial PE, and is beneficial to the proposed \texttt{LSPE} architecture as demonstrated by the gain of performance in several experiments.

\subsection{Models used for comparison in Table \ref{tab:comparison_sota}}
\label{sec:models_comparison_sota}
As a complete reference, the different GNN baselines and SOTA models that are used for the comparison in Table \ref{tab:comparison_sota} are Graph Convolutional Networks (GCN) \citep{kipf2017semi}, Graph Attention Networks (GAT) \citep{velivckovic2018graph}, GatedGCN-LapPE \citep{bresson2017residual, dwivedi2020benchmarking}, Graph Transformer (GT) \citep{dwivedi2021generalization}, Spectral Attention Networks (SAN) \citep{kreuzer2021rethinking}, Graphormer \citep{ying2021transformers}, Graph Isomorphism Networks (GIN) \citep{xu2018how}, DeeperGCN \citep{li2020deepergcn}, Principle Neighborhood Aggregation (PNA) \citep{corso2020principal}, Directional Graph Networks (DGN) \citep{beani2021directional} and Parameterized Hypercomplex GNNs (PHC-GNN) \citep{le2021parameterized}.

\subsection{Figure for the study of $k$ steps in RWPE (Section \ref{sec:experiments_ablation_studies})}
\label{sec:figure_k_study}

\begin{figure}[h]
\centering
  \begin{subfigure}{0.42\linewidth}
  \centering
    \includegraphics[width=\linewidth]{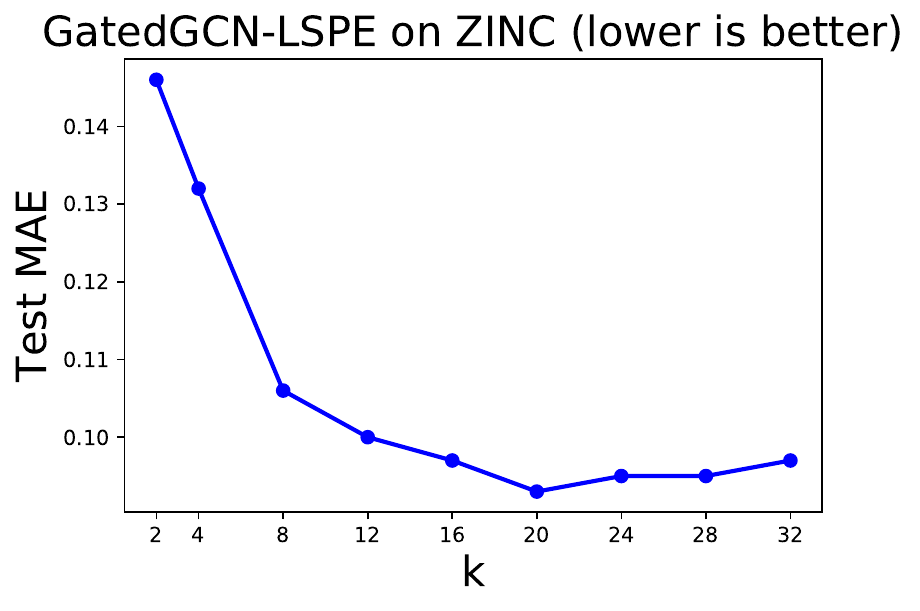}
    \label{fig:k_plot_ZINC}
  \end{subfigure}
  \begin{subfigure}{0.40\linewidth}
  \centering
    \includegraphics[width=\linewidth]{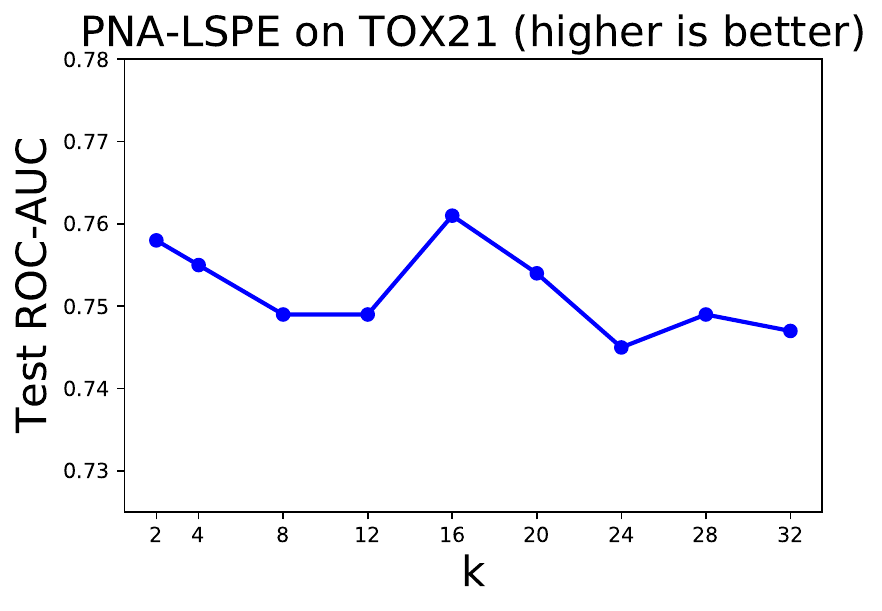}
    \label{fig:k_plot_TOX21}
  \end{subfigure}
  \vspace{-16pt}
  \caption{Test scores on selecting different values of $k$ which is used to determine the number of iterative steps of RW in RWPE as well as the dimension of the learned PE at the final layer, Eqn. \ref{eqn:final_loss}.}
  \label{fig:k_plots}
  \vspace{-12pt}
\end{figure}

\section{Related work in Detail}
\label{sec:related_work_detailed}

In this detailed section on related work, we first review the limitations of existing MP-GNN architectures in terms of their theoretical expressiveness, suggesting possible improvements to make GNNs more powerful. Then, we introduce a number of  non-learned and learning techniques that can be studied under the umbrella of graph positional encoding. Finally, we highlight the recent developments for generalizing Transformers to graphs. Our aim is to connect meaningful innovations through the detailed background on these three research directions, the unification of which spearheaded the development of this work.

\subsection{Theoretical Expressivity and Weisfeiler-Leman GNNs}
\label{sec:related_work_theory}
{\bf Weisfeiler-Leman test.} The limitation of MP-GNNs in failing to distinguish non-isomorphic graphs was first carefully studied in \cite{xu2018how} and \cite{morris2019weisfeiler}, based on the equivalence of MP-GNNs and the 1-WL isomorphism test \citep{weisfeiler1968reduction}. As such, MP-GNNs may perform poorly on graphs that exhibit several symmetries in their original structure, such as node and edge isomorphisms \citep{murphy2019relational, srinivasan2019equivalence}. Besides, some message-passing functions may not be discriminative enough \citep{xu2018how,corso2020principal}.

{\bf Equivariant GNNs.} Graph Isomorphism Networks (GINs) \citep{xu2018how} were designed to be as maximally expressive as the original 1-WL test \citep{weisfeiler1968reduction}. However, the 1-WL test can fail to distinguish (simple) non-isomorphic graphs, thus requiring novel GNNs with more expressivity power. As the original 1-WL test only considers 2-tuple of nodes, i.e. the standard edges in a graph, a natural approach to improve the expressivity power of the 1-WL test is to examine higher-order interactions between nodes with $k$-tuple of nodes with $k\geq 3$. To this end, $k$-order Equivariant-GNNs were introduced in \cite{maron2018invariant}. But these architectures require $O(n^k)$ memory and speed complexities. This is an important practical limitation as $k=3$ is at least needed to design more powerful GNNs than GINs. Along this line, the most efficient WL-GNNs that have been proposed are in \cite{maron2019provably,chen2019equivalence,azizian2020expressive}, which have $O(n^2)$ memory and $O(n^3)$ speed complexities.

\subsection{Graph Positional Encoding}
\label{sec:related_work_graphpe}

{\bf Importance of Positional Information.} The idea of positional encoding, i.e. the notion of global position of pixels in images, words in texts and nodes in graphs, plays a central role in the effectiveness of the most prominent neural networks with ConvNets \citep{lecun1998gradient}, RNNs \citep{hochreiter1997long}, and Transformers \citep{vaswani2017attention}. These architectures integrate structural and positional attributes of data when building abstract feature representations. For instances, ConvNets intrinsically consider regular spatial structure for the position of pixels \citep{Islam2020How}, RNNs also build on the sequential structure of the word positions, and Transformers employ positional encoding of words (see \cite{dufter2021position} for a review). For GNNs, the position of nodes is more challenging due to the fact that there does not exist a canonical positioning of nodes in arbitrary graphs. This implies that there is no obvious notion of global and relative position of nodes, and consequently no specific directions on graphs (like the top, down, left and right directions in images). Despite these issues, graph positional encoding are as much critical for GNNs as they are for ConvNets, RNNs and Transformers, as demonstrated for prediction tasks on graphs \citep{srinivasan2019equivalence, cui2021positional}.

{\bf Index Positional Encoding.} \cite{Loukas2020What} identified another cause of the limited expressivity of the standard MP-GNNs. Such GNNs do not have the capacity to handle anonymous nodes, i.e. nodes which do not have unique node features. This property turns out to be critical to show that MP-GNNs can be universal approximators if each node in the graph can be assigned to a unique or discriminating feature. The theorem results from an alignment between MP-GNNs and distributed local algorithms \citep{naor1995can, sato2019approximation}. In order to address the issue of anonymous MP-GNNs and improve their theoretical expressiveness w.r.t the WL test, \cite{murphy2019relational} introduced Graph Relational Pooling. Their model assigns a unique identifier to each node, defined by an indexing of the nodes. However, such a model must be trained with the $n!$ possible index permutations to guarantee higher expressivity, which is not computationally feasible. As a consequence, during training, node indexing is uniformly sampled from the $n!$ possible choices in order for their network to learn to be independent to the choice of the index PE at test time. Similarly, random node identifier could be used for breaking the node anonymity. Yet, this PE also suffers from the lack of generalization for unseen graphs \citep{Loukas2020What}.

{\bf Laplacian Positional Encoding.} Besides providing a unique representation for each node, meaningful graph PE should also be  permutation-invariant and distance-sensitive, meaning that the difference between the PEs of two nodes far apart on the graph must be large, and small for two nodes nearby. Laplacian eigenvectors \citep{belkin2003laplacian} appear to be good candidates for graph PE, belonging to the class of unsupervised manifold learning techniques. Precisely, they are spectral techniques that embed graphs into an Euclidean space, and are defined via the factorization of the graph Laplacian $\Delta=\textrm{I}_n-D^{-1/2}AD^{-1/2}=U^T\Lambda U$,  
where $\textrm{I}_n$ is the $n\times n$ identity matrix, $A$ the $n\times n$ adjacency matrix, $D$ the $n\times n$ degree matrix, and $n\times n$ matrices $\Lambda$ and $U$ correspond to the eigenvalues and eigenvectors respectively. The complexity for computing this full factorization is $O(E^{3/2})$ and $O(n)$ with approximate Nystrom method \citep{fowlkes2004spectral}. Laplacian eigenvectors form a meaningful local coordinate system, while preserving the global graph structure. As these eigenvectors hold the key properties of permutation-invariant, uniqueness, computational efficiency and distance-aware w.r.t. the graph topology, they were proposed as graph PE \citep{dwivedi2020benchmarking,dwivedi2021generalization}. They also naturally generalize the positional encoding used in  Transformers \citep{vaswani2017attention} to arbitrary graphs. The main limitation of this graph PE is the existence of a sign ambiguity as eigenvectors are defined up to $\pm 1$. This leads to $2^k$ number of possible sign values when selecting $k$ number of eigenvectors. In practice, we choose $k\leq n$ eigenvectors given the manifold assumption, and therefore $2^k$ is much smaller $n!$ (the number of possible ordering of the nodes), and therefore smaller amount of ambiguities to be resolved by the network. During the training, eigenvectors are uniformly sampled at random between the $2^k$ possibilities \citep{dwivedi2020benchmarking, kreuzer2021rethinking} 
in order for the network to learn to be invariant w.r.t the sign of the eigenvectors.

{\bf Other graph PE.} \cite{li2020distance} proposed the use of distance encoding (DE) as node attributes, and additionally as controller of message aggregation. DE captures relative distances between nodes in a graph using powers of the random walk matrix. The resulting GNN was shown to have better expressivity than the 1-WL test. However, the limitation on regular graphs, and the cost and memory requirement of using power matrices may prevent the use of this technique to larger graphs. \cite{Khasahmadi2020Memory-Based} used random walk with restart \citep{pan2004automatic} as topological embeddings with the initial node features.

\cite{you2019position} proposed learnable position-aware embeddings based on random anchor sets of nodes for pairwise nodes (or link) tasks. This work also seeks to develop positional encoding that can be learned along with the structural representation within the GNN. However, the random selection of anchors  has its limitations, which makes their approach less generalizable on inductive tasks.

\cite{bouritsas2020improving,bodnar2021weisfeiler} introduced hybrid GNNs based on the WL-test and the message-passing 
aggregation
mechanism. These networks use prior knowledge about a class of graphs of interest such as rings for molecules and cliques for social networks. The prior information is then encoded into MP-GNNs to obtain more expressive models by showing that the such GNNs are not less powerful than the 3-WL test. They obtained top performance on molecular datasets but the prior information regarding graph sub-structures needs to be pre-computed, and sub-graph matching and counting require $O(n^k)$ for $k$-tuple sub-structure. Besides, complexity of the message passing process depends linearly w.r.t. the size of the sub-graph structure. Note that the core idea of substructure counting with e.g. the number of rings associated to an atom provides a powerful higher-order structural information to the network and can improve significantly the tasks related to substructure counting.

\subsection{Transformer-based GNNs}
\label{sec:related_work_transfGNNs}

MP-GNNs are GNNs that leverage the sparse graph structure as computational graph, allowing training and inference with linear complexity and making them scalable to medium and large-scale graphs. However, besides their low expressivity, these GNNs hold two important and well-identified limitations. Firstly, MP-GNNs are susceptible to the information bottleneck limitation a.k.a. over-squashing \citep{alon2020bottleneck} when messages from across distant nodes are aggregated to a node. Secondly, long-range interactions between far away nodes can also be limited, and require multiple layers that can suffer from the vanishing gradient problem. These limitations are similar to the ones present in Recurrent Neural Networks (RNNs) \citep{hochreiter1997long}, and can lead MP-GNNs to perform poorly on tasks where long-range interactions are necessary.

To overcome these limitations, it seems natural to use Transformer networks \citep{vaswani2017attention} which alleviates the long-range issue as `everything is connected to everything'. However, it was found that the direct adoption of full-graph operable Transformers perform poorly compared to MP-GNNs on graph structured datasets \citep{dwivedi2021generalization}. Besides, Transformer-based GNNs require to replace $O(n)$ complexity with $O(n^2)$. So these GNNs are limited to small graphs like molecules and cannot scale to larger ones like social graphs or knowledge graphs. \cite{dwivedi2021generalization} designed a sparsely-connected architecture called GraphTransformer that reduces the complexity to $O(E)$ by considering the graph topology instead of connecting each node to all other nodes, similar to GATs \citep{velivckovic2018graph}. Still, the GraphTransformer was unable to outperform SOTA GNNs on benchmark datasets. 
Along this line, \cite{kreuzer2021rethinking} recently proposed Spectral Attention Networks (SANs), a fully-graph operable Transformer model that improves GraphTransformer \citep{dwivedi2021generalization} with two contributions. First, the authors designed a learnable PE module based on self-attention 
applied to the Laplacian eigenvectors, and injected this resultant PE into the input layer of the network. Second, SANs separated the parameters for real edges and complementary (non-real) edges, enabling the model to process the available sparse graph structure and long-range node connections in a learnable manner. However, their learned PE, based on the Laplacian eigenvectors, inherently exhibits the limitation of sign ambiguity. \cite{kreuzer2021rethinking} attempted at alleviating the sign ambiguity through another architecture named Edge-Wise LPE. However, the architecture's complexity being $O(n^4)$ makes it a practically infeasible model.
GraphiT \citep{mialon2021graphit} and Graphormer \citep{ying2021transformers} were also very recently developed as full-graph operable Transformers for graphs with the idea to weigh (or, control) the attention mechanism based on the graph topology. Specifically, GraphiT employs diffusion geometry to capture short-range and long-range graph information, and Graphormer uses shortest paths. Altogether, these works exploit different relative positional encoding information to improve the expressivity of Transformers for graphs.

\section{Instances of \texttt{LSPE} with Sparse and Transformer GNNs}
\label{sec:model_eqns_supplementary}

\subsection{Sparse GNNs with \texttt{LSPE}}
\label{sec:sparse_gnns_lspe}
In this section, we augment two MP-GNN architectures with learnable positional representation, namely GatedGCN \citep{bresson2017residual} and PNA \citep{corso2020principal}.

\subsubsection{GatedGCN\texttt{-LSPE}}
\label{sec:gatedgcn_lspe}
GatedGCNs \citep{bresson2017residual} are sparse MP-GNNs equipped with a soft-attention mechanism that is able to learn adaptive edge gates to improve the message aggregation step of GCN networks \citep{kipf2017semi}.  We augment this model to develop GatedGCN\texttt{-LSPE}, defined as:
\begin{align}
    h^{\ell+1}, e^{\ell+1}, p^{\ell+1} &= \text{GatedGCN}\texttt{-LSPE}\Big(h^{\ell}, e^{\ell}, p^{\ell}\Big), \ h\in\mathbb{R}^{n\times d}, e\in\mathbb{R}^{E\times d}, p\in\mathbb{R}^{n\times d},\\
    \text{with} \ \ h_i^{\ell+1} &= h_i^{\ell} + \text{ReLU}\Big( \text{BN} \Big( A_1^{\ell} \left[ \!\!\begin{array}{c} h_i^\ell \\ p_i^\ell \\ \end{array} \!\!\right] + \sum_{j \in \mathcal{N}(i)} \eta_{ij}^{\ell} \odot A_2^{\ell} \left[ \!\!\begin{array}{c} h_j^\ell \\ p_j^\ell \\ \end{array} \!\!\right] \Big) \Big), \label{eqn:gatedcn_lspe_1}\\
e_{ij}^{\ell+1} &= e_{ij}^{\ell} + \text{ReLU}\big(\text{BN}\big(\hat{\eta}_{ij}^{\ell}\big)\big),\label{eqn:gatedgcn_lspe_2}\\
p_i^{\ell+1} &= p_i^{\ell} + \tanh \Big( C_1^{\ell}p_i^{\ell} + \sum_{j \in \mathcal{N}(i)} \eta_{ij}^{\ell} \odot C_2^{\ell}
  p_j^{\ell} \Big), \label{eqn:gatedgcn_lspe_3}\\
\text{and} \ \ \eta_{ij}^{\ell} &= \frac{\sigma\big(\hat{\eta}_{ij}^{\ell}\big)}{ \sum_{j' \in \mathcal{N}(i)} \sigma\big(\hat{\eta}_{ij'}^{\ell}\big) + \epsilon},\\[4pt]
\hat{\eta}_{ij}^{\ell} &= B_1^{\ell}h_i^{\ell} + B_2^{\ell}h_j^{\ell} + B_3^{\ell}e_{ij}^{\ell},
\end{align}
where $h_i^{\ell}, e_{ij}^{\ell}, p_i^{\ell}, {\eta}_{ij}^{\ell}, \hat{\eta}_{ij}^{\ell} \in \mathbb{R}^{d}, A_1^{\ell}, A_2^{\ell} \in \mathbb{R}^{d \times 2d}$ and $B_1^{\ell}, B_2^{\ell}, B_3^{\ell}, C_1^{\ell}, C_2^{\ell} \in \mathbb{R}^{d \times d}$.

\subsubsection{PNA\texttt{-LSPE}}
\label{sec:pna_lspe}
PNA \citep{corso2020principal} is a sparse MP-GNN model which uses a combination of node aggregators to overcome the theoretical limitation of a single aggregator. We propose PNA\texttt{-LSPE} whose layer update equation is defined as:
\begin{align}
h^{\ell+1}, p^{\ell+1} &= \text{PNA}\texttt{-LSPE}\Big(h^{\ell}, e^0, p^{\ell}\Big), \ h\in\mathbb{R}^{n\times d}, e^0\in\mathbb{R}^{E\times d}, p\in\mathbb{R}^{n\times d},\\
\text{with} \ \ h_{i}^{\ell+1} &= h_{i}^{\ell} + \text{LReLU}\Big( \text{BN} \Big(U_h^{\ell}\left(\left[ \!\!\begin{array}{c} h_i^\ell \\ p_i^\ell \\ \end{array} \!\!\right], \bigoplus_{j \in \mathcal{N}(i)} M_h^{\ell}\left(\left[ \!\!\begin{array}{c} h_i^\ell \\ p_i^\ell \\ \end{array} \!\!\right], e_{ij}^0, \left[ \!\!\begin{array}{c} h_j^\ell \\ p_j^\ell \\ \end{array} \!\!\right]\right)\right)\Big)\Big), \label{eqn:pna_lspe_1}\\
p_{i}^{\ell+1} &= p_{i}^{\ell} + \tanh \Big(U_p^{\ell}\left(p_i^{\ell}, \bigoplus_{j \in \mathcal{N}(i)} M_p^{\ell}\Big(p_i^{\ell}, e_{ij}^0, p_j^{\ell}\Big)\right)\Big), \label{eqn:pna_lspe_2}\\
\text{and} \ \ \bigoplus \ &= \ \ {\left[\begin{array}{c}
I \\
S(D, \alpha=1) \\
S(D, \alpha=-1)
\end{array}\right]} \ \ \otimes \ \ {\left[\begin{array}{c}
\mu \\
\sigma \\
\max \\
\min
\end{array}\right]}, \label{eqn:pna_layer_agg_scalers}
\end{align} 
where $\bigoplus$ is the principal aggregator designed in \citep{corso2020principal}, LReLU stands for LeakyReLU activation,
amd $U_h^{\ell}, U_p^{\ell}, M_h^{\ell}$ and $M_p^{\ell}$ are linear layers (or multi-layer perceptrons) with learnable parameters.

\subsection{Transformer GNNs with \texttt{LSPE}}
\label{sec:transformer_lspe}

The recently developed SAN \citep{kreuzer2021rethinking}, GraphiT \citep{mialon2021graphit} and Graphormer \citep{ying2021transformers} are promising full-graph operable Transformers incorporating several methods to encode positional and structural features into the network. In the next sections, we expand these Transformer-based networks with the proposed \texttt{LSPE} architecture.

\subsubsection{SAN\texttt{-LSPE}}
\label{sec:san_lspe}
Like Transformers, Spectral Attention Networks (SAN) \citep{kreuzer2021rethinking} operate on full graphs although the network separates the parameters coming from existing edges and non-existing edges in the graph. Furthemore, the contribution of attentions from existing and non-existing edges are weighted by an additive positive scalar $\gamma$, which can be tuned for different tasks. SAN also considers a Learnable Positional Encoding (LPE) module which takes in Laplacian eigenvectors and transforms them into a fixed size PE with a self-attention encoder. This PE is then used in the main architecture in a manner similar to MP-GNNs\texttt{-PE} as defined in Eq. (\ref{eqn:mpgnn-pe-input}). We propose to extend SAN by replacing the LPE module with the \texttt{LSPE} architecture proposed in Section \ref{sec:generic_formulation} where positional representation is learned in line with structural embedding at each GNN layer:
\begin{align}
h^{\ell+1}, p^{\ell+1} &= \text{SAN}\texttt{-LSPE}\Big(h^{\ell}, e^0, p^{\ell}\Big), \ h\in\mathbb{R}^{n\times d}, e^0\in\mathbb{R}^{n\times n\times d}, p\in\mathbb{R}^{n\times d},\label{eqn:san_lspe_1}\\
\text{with} \ \ h_{i}^{\ell+1}&=\text{BN}\left(\bar{h}_{i}^{\ell+1}+  W_{2}^{\ell}\ \text{ReLU}\left( W_{1}^{\ell}\ \bar{h}_{i}^{\ell+1} \right) \right)\in\mathbb{R}^{d}\\
\bar{h}_{i}^{\ell+1}&=\text{BN}\left(h_{i}^{\ell}+\hat{h}_{i}^{\ell+1}\right)\in\mathbb{R}^{d},\label{eqn:san_lspe_2}
\end{align}
\begin{eqnarray}
\hat{{h}}_{i}^{\ell+1}= O^{\ell}\Big(\bigparallel_{k=1}^{H}\sum_{j \in \mathcal{V}} \frac{w_{ij}^{k, \ell}}{\sum_{j' \in \mathcal{V}} w_{ij'}^{k, \ell}} \ v_j^{k,\ell} \Big)\in\mathbb{R}^{d},\label{eqn:san_lspe_3}
\end{eqnarray}
\begin{eqnarray}
w_{ij}^{k,\ell}=\left\{\begin{array}{ll}
\frac{1}{1+\gamma} \cdot \text{exp}(A_{ij}^{k,\ell}) & \textrm{ if } ij\in E  \\
\frac{\gamma}{1+\gamma} \cdot \text{exp}(\bar{A}_{ij}^{k,\ell}) & \textrm{ if } ij\not\in E \\
\end{array}\right.,\label{eqn:san_lspe_3a}
\end{eqnarray}
\begin{eqnarray}
\left\{
\begin{array}{ll}
A_{ij}^{k,\ell}={q_i^{k,\ell}}^T \textrm{diag}(c_{ij}^{k,\ell}) k_j^{k,\ell} / \sqrt{d_k}\in\mathbb{R} & \textrm{ if } ij\in E \\
\bar{A}_{ij}^{k,\ell}=\bar{q}_i{{}^{k,\ell}}^T\textrm{diag}({\bar{c}}_{ij}^{k,\ell}) {\bar{k}}_j^{k,\ell} / \sqrt{d_k}\in\mathbb{R} & \textrm{ if } ij\not\in E 
\end{array}
\right.
\end{eqnarray}
\begin{eqnarray}
Q^{k,\ell} &\!=\! \left[ \!\!\begin{array}{c} h^\ell \\ p^\ell \\ \end{array} \!\!\right] W_{Q}^{k,\ell},\
K^{k,\ell} \!=\! \left[ \!\!\begin{array}{c} h^\ell \\ p^\ell \\ \end{array} \!\!\right] W_{K}^{k,\ell},\
V^{k,\ell} \!=\! \left[ \!\!\begin{array}{c} h^\ell \\ p^\ell \\ \end{array} \!\!\right] W_{V}^{k,\ell}\in\mathbb{R}^{n\times d_k}\\
\bar{Q}^{k,\ell} &\!=\! \left[ \!\!\begin{array}{c} h^\ell \\ p^\ell \\ \end{array} \!\!\right] \bar{W}_{Q}^{k,\ell},\
\bar{K}^{k,\ell} \!=\! \left[ \!\!\begin{array}{c} h^\ell \\ p^\ell \\ \end{array} \!\!\right] \bar{W}_{K}^{k,\ell},\
\bar{V}^{k,\ell} \!=\! \left[ \!\!\begin{array}{c} h^\ell \\ p^\ell \\ \end{array} \!\!\right] \bar{W}_{V}^{k,\ell}\in\mathbb{R}^{n\times d_k}
\end{eqnarray}
\begin{eqnarray}
C^{k,0} \!=\! e^0 W_{e}^{k}, \ \bar{C}^{k,0} \!=\! e^0 \bar{W}_{e}^{k}\in\mathbb{R}^{n\times n\times d_k} 
\end{eqnarray}
\begin{align}
\text{and} \ \ {p}_{i}^{\ell+1}&= p_i^{\ell} + \tanh \Big( {O}_{p}^{\ell} \Big( \bigparallel_{k=1}^{H}\sum_{j \in \mathcal{V}} \frac{w_{p,i j}^{k, \ell}}{\sum_{j' \in \mathcal{V}} w_{p,i j'}^{k, \ell}} v_{p,j}^{k,\ell}\Big)\Big)\in\mathbb{R}^d,\label{eqn:san_lspe_5}
\end{align}
\begin{eqnarray}
w_{p,ij}^{k,\ell}=\left\{\begin{array}{ll}
\frac{1}{1+\gamma} \cdot \text{exp}(A_{p,ij}^{k,\ell}) & \textrm{ if } ij\in E  \\
\frac{\gamma}{1+\gamma} \cdot \text{exp}(\bar{A}_{p,ij}^{k,\ell}) & \textrm{ if } ij\not\in E \\
\end{array}\right.,\label{eqn:san_lspe_3a_p}
\end{eqnarray}
\begin{eqnarray}
\left\{
\begin{array}{ll}
A_{p,ij}^{k,\ell}={q_{p,i}^{k,\ell}}^T \textrm{diag}(c_{p,ij}^{k,\ell}) k_{p,j}^{k,\ell} / \sqrt{d_k}\in\mathbb{R} & \textrm{ if } ij\in E \\
\bar{A}_{p,ij}^{k,\ell}=\bar{q}_{p,i}{{}^{k,\ell}}^T\textrm{diag}({\bar{c}}_{p,ij}^{k,\ell}) {\bar{k}}_{p,j}^{k,\ell} / \sqrt{d_k}\in\mathbb{R} & \textrm{ if } ij\not\in E 
\end{array}
\right.
\end{eqnarray}
\begin{eqnarray}
Q_p^{k,\ell} &\!=\! p^\ell W_{p,Q}^{k,\ell},\
K_p^{k,\ell} \!=\! p^\ell W_{p,K}^{k,\ell},\
V_p^{k,\ell} \!=\! p^\ell W_{p,V}^{k,\ell}\in\mathbb{R}^{n\times d_k}\\
\bar{Q}_p^{k,\ell} &\!=\! p^\ell \bar{W}_{p,Q}^{k,\ell},\
\bar{K}_p^{k,\ell} \!=\! p^\ell \bar{W}_{p,K}^{k,\ell},\
\bar{V}_p^{k,\ell} \!=\! p^\ell \bar{W}_{p,V}^{k,\ell}\in\mathbb{R}^{n\times d_k}
\end{eqnarray}
\begin{eqnarray}
C_p^{k,0} \!=\! e^0 W_{p,e}^{k}, \ \bar{C}_p^{k,0} \!=\! e^0 \bar{W}_{p,e}^{k}\in\mathbb{R}^{n\times n\times d_k} 
\end{eqnarray}
where $W_1^{\ell}, W_2^{\ell} \in \mathbb{R}^{d \times d}$, $O^{\ell}, O_p^{\ell} \in \mathbb{R}^{d \times d}$, $W_Q^{k,\ell},W_K^{k,\ell},W_V^{k,\ell},\bar{W}_Q^{k,\ell},\bar{W}_K^{k,\ell},\bar{W}_V^{k,\ell} \in \mathbb{R}^{2d \times d_k}$, $W_{p,Q}^{k,\ell},W_{p,K}^{k,\ell},W_{p,V}^{k,\ell},\bar{W}_{p,Q}^{k,\ell},\bar{W}_{p,K}^{k,\ell},\bar{W}_{p,V}^{k,\ell} \in \mathbb{R}^{d \times d_k}$, $W_{e}^{k}, \bar{W}_{e}^{k}, W_{p,e}^{k}, \bar{W}_{p,e}^{k}\in \mathbb{R}^{d \times d_k}$, and $d_k = d/H$ is the dimension of the $k^{th}$ head for a total of $H$ heads. BN denotes the standard Batch Normalization \citep{ioffe2015batch}. Finally, we make the balance scalar parameter $\gamma\geq 0$ learnable (also clipping its range in $[0,1]$) differently from \citep{kreuzer2021rethinking} where its optimal value is computed by grid search.

\subsubsection{GraphiT\texttt{-LSPE}}
\label{sec:graphit_lspe}

Similarly to SAN, GraphiT \citep{mialon2021graphit} is a full-graph operable Transformer GNN which makes use of the diffusion distance to capture short-range and long-range interactions between nodes depending of the graph topology. This pairwise diffusion distance is used as a multiplicative weight to adapt the weight scores to the closeness or farness of the nodes. For example, if two nodes are close on the graph, them the diffusion distance $K_{ij}$ will have a value close to one, and when the two nodes are far away then the value of $K_{ij}$ will be small. 

Unlike SAN, the GraphiT model does not consider separate parameters for existing and non-existing edges for a graph. However, following \cite{kreuzer2021rethinking} and our experiments,  separating the parameters for each type of edges showed to improve the performance. Therefore, we augment the original GraphiT architecture with learnable positional features and use two sets of parameters for the edges and the complementary edges to define GraphiT\texttt{-LSPE}. The GraphiT\texttt{-LSPE} model uses the same update equation as SAN\texttt{-LSPE} except for the weight score which are re-defined to introduce the diffusion kernel:
\begin{eqnarray}
w_{ij}^{k,\ell}=\left\{\begin{array}{ll}
K_{ij} \cdot \text{exp}(A_{ij}^{k,\ell}) & \textrm{ if } ij\in E  \\
K_{ij} \cdot \text{exp}(\bar{A}_{ij}^{k,\ell}) & \textrm{ if } ij\not\in E \\
\end{array}\right..\label{eqn:san_lspe_3a_graphit}
\end{eqnarray}
Following \citep{mialon2021graphit}, the diffusion distance is chosen to be the $p$-step random walk kernel defined as $K = (\textrm{I}_n - \beta \Delta)^p \in \mathbb{R}^{n \times n}$ where $\textrm{I}_n, \Delta \in \mathbb{R}^{n \times n}$ is the identity matrix and the graph Laplacian matrix respectively. Hyper-parameter $\beta$ controls the amount of diffusion with value between $[0.25,0.50]$.

\section{Experiments on non-molecular graphs}
\label{sec:experiments_non_molecular}

\begin{table}[!htb]
    \centering
    \caption{Results on the IMDB-MULTI, IMDB-BINARY and CIFAR10 superpixels. All scores are averaged over 4 runs with 4 different seeds. On IMDB- each seed experiment is on 10-fold cross validation. \textbf{Bold}: GNN's best score. No PosLoss is used with \texttt{LSPE}. $\dagger$ denotes the result is taken directly from \citep{dwivedi2020benchmarking}.}
    \scalebox{0.74}{
    \begin{tabular}{rrrc|cccccc}
        \toprule
        \textbf{Dataset} & \textbf{Model} &  \textbf{Init PE} & \textbf{\texttt{LSPE}} & \textbf{$L$} & \textbf{\#Param} & \textbf{TestAcc$\pm$s.d.} & \textbf{TrainAcc$\pm$s.d.} & \textbf{Epochs} & \textbf{Epoch/Total} \\
        \midrule
        \multirow{2}{*}{IMDB-B} & GatedGCN & \textbf{x} & \textbf{x} & 4 & 87122 & 66.050$\pm$6.631 & 67.769$\pm$4.675 & 115.85 & 0.45s/0.15hr\\
        & GatedGCN & \textbf{RWPE} & \textbf{\checkmark} & 4 & 91470 & \textbf{70.025$\pm$5.147} & 72.166$\pm$1.706 & 111.17 & 0.56s/0.19hr\\
        \midrule
        \multirow{2}{*}{IMDB-M} & GatedGCN & \textbf{x} & \textbf{x} & 4 & 87139 & 45.767$\pm$4.906 & 47.725$\pm$1.803 & 109.18 & 0.55s/0.17hr\\
        & GatedGCN & \textbf{RWPE} & \textbf{\checkmark} & 4 & 91483 & \textbf{46.467$\pm$3.997} & 48.781$\pm$1.568 & 100.55 & 0.72s/0.22hr\\
        \midrule
        \multirow{2}{*}{CIFAR10} & GatedGCN & \textbf{x} & \textbf{x} & 4 & 104357 & 67.312$\pm$0.311 & 94.553$\pm$1.018 & 97.00 & 154.15s/4.22hr $\dagger$\\
        & GatedGCN & \textbf{RWPE} & \textbf{\checkmark} & 4 & 107237 & \textbf{70.858$\pm$0.631} & 78.616$\pm$1.006 & 185.00 & 68.51s/3.89hr\\
        \bottomrule

    \end{tabular}
    }
    \label{tab:lspe_non_molecular_graphs}
\end{table}

We conduct experiments on 3 non-molecular graph datasets to demonstrate the effectiveness of the proposed \texttt{LSPE} architecture on any graph domain in general. We select GatedGCN as the GNN instance here. The datasets used are from the domains of social network (IMDB-BINARY and IMDB-MULTI \citep{morris2020tudataset}) and image superpixels (CIFAR10 \citep{dwivedi2020benchmarking}) with graph classification being the prediction task. 

IDBM-BINARY and IMDB-MULTI contain 1,000 and 1,500 graphs respectively which are ego-networks extracted from actor collaboration graphs. There are 2 classes in IMDB-BINARY and 3 classes in IMDB-MULTI with the class denoting the genre of the graph. CIFAR10 is a superpixel dataset of 60,000 graphs where each graph represents a connectivity structure of the image superpixels as nodes. There are 10 classes to be predicted as with the original CIFAR10 image dataset \citep{krizhevsky2009learning}.

In Table \ref{tab:lspe_non_molecular_graphs}, we show the advantage of using \texttt{LSPE} by instantiating GatedGCN\texttt{-LSPE} to train and evaluate on these non-molecular graphs. For evaluation and reporting of results, we follow the respective protocols as specified in \cite{morris2020tudataset, dwivedi2020benchmarking} while comparing two models on a dataset on the similar range of model parameters. In accordance with the molecular datasets (Section \ref{sec:experiments_results}), we consistently observe performance gains on each of the three datasets in Table \ref{tab:lspe_non_molecular_graphs}. This result further justifies the usefulness of \texttt{LSPE} to be applicable in general for representation learning on any graph domain.

\section{Additional model configuration details}
\label{sec:supplementary_implementation}

\begin{table}[!htb]
    \centering
    \caption{Additional hyperparamters for the models used in Table \ref{tab:all_lspe_expts}. \textbf{$k$} is the dimension of PE, or the steps of random walk if the PE is RWPE. \textbf{$\beta$} and \textbf{$p$} is applicable to GraphiT (Sec. \ref{sec:graphit_lspe}). \textbf{Init\_lr} and \textbf{Min\_lr} are the initial and final learning rates for the learning rate decay strategy where the lr decays with a reduce \textbf{Factor} if the validation score doesn't improve after the \textbf{Patience} number of epochs. \textbf{$\alpha$} and \textbf{$\lambda$} are applicable when PosLoss is used (Eqn. \ref{eqn:final_loss}).}
    \scalebox{0.74}{
    \begin{tabular}{lrrcc|ccc|cccc|cc}
        \toprule
        \parbox[t]{2mm}{\multirow{14}{*}{\rotatebox[origin=c]{90}{ZINC}}} & \textbf{Model} &  \textbf{Init PE} & \textbf{\texttt{LSPE}} & \textbf{PosLoss} & \textbf{$k$} & \textbf{$\beta$} & \textbf{$p$} & \textbf{Init\_lr} & \textbf{Patience} & \textbf{Factor} & \textbf{Min\_lr} & \textbf{$\alpha$} & \textbf{$\lambda$}\\
        \midrule
        & GatedGCN & \textbf{x} & \textbf{x} &\textbf{x} & - & - & - & 1e-3 & 25 & 0.5 & 1e-6 & - & -\\
        & GatedGCN & \textbf{LapPE} & \textbf{x} & \textbf{x} & 8 & - & - & 1e-3 & 25 & 0.5 & 1e-6 & - & -\\
        & GatedGCN & \textbf{RWPE} & \textbf{\checkmark} & \textbf{x} & 20 & - & - & 1e-3 & 25 & 0.5 & 1e-6 & - & -\\
        & GatedGCN & \textbf{RWPE} & \textbf{\checkmark} & \textbf{\checkmark} & 20 & - & - & 1e-3 & 25 & 0.5 & 1e-6 & 1 & 1e-1\\
        \cmidrule{2-14}
        & PNA & \textbf{x} & \textbf{x} & \textbf{x} & - & - & - & 1e-3 & 25 & 0.5 & 1e-6 & - & -\\
        & PNA & \textbf{RWPE} & \textbf{\checkmark} & \textbf{x} & 16 & - & - & 1e-3 & 25 & 0.5 & 1e-6 & - & - \\
        \cmidrule{2-14}
        & SAN & \textbf{x} & \textbf{x} & \textbf{x} & - & - & - & 3e-4 & 25 & 0.5 & 1e-6 & - & -\\
        & SAN & \textbf{RWPE} & \textbf{\checkmark} & \textbf{x} & 16 & - & - & 7e-4 & 25 & 0.5 & 1e-6 & - & -\\
        \cmidrule{2-14}
        & GraphiT & \textbf{x} & \textbf{x} & \textbf{x} & - & 0.25 & 16 & 3e-4 & 25 & 0.5 & 1e-6 & - & -\\
        & GraphiT & \textbf{RWPE} & \textbf{\checkmark} & \textbf{x} & 16 & 0.25 & 16 & 7e-4 & 25 & 0.5 & 1e-6 & - & -\\
        \toprule
        \parbox[t]{2mm}{\multirow{14}{*}{\rotatebox[origin=c]{90}{MOLTOX21}}} & \textbf{Model} &  \textbf{Init PE} & \textbf{\texttt{LSPE}} & \textbf{PosLoss} & \textbf{$k$} & \textbf{$\beta$} & \textbf{$p$} & \textbf{Init\_lr} & \textbf{Patience} & \textbf{Factor} & \textbf{Min\_lr} & \textbf{$\alpha$} & \textbf{$\lambda$} \\
        \midrule
        & GatedGCN & \textbf{x} & \textbf{x} & \textbf{x} & - & - & - & 1e-3 & 25 & 0.5 & 1e-5 & - & - \\
        & GatedGCN & \textbf{LapPE} & \textbf{x} & \textbf{x} & 3 & - & - & 1e-3 & 25 & 0.5 & 1e-5 & - & - \\
        & GatedGCN & \textbf{RWPE} & \textbf{\checkmark} & \textbf{x} & 16 & - & - & 1e-3 & 25 & 0.5 & 1e-5 & - & - \\
        \cmidrule{2-14}
        & PNA & \textbf{x} & \textbf{x} & \textbf{x} & - & - & - & 5e-4 & 10 & 0.8 & 2e-5 & - & - \\
        & PNA & \textbf{RWPE} & \textbf{\checkmark} & \textbf{x} & 16 & - & - & 5e-4 & 10 & 0.8 & 2e-5 & - & -  \\
        & PNA & \textbf{RWPE} & \textbf{\checkmark} & \textbf{\checkmark} & 16 & - & - & 5e-4 & 10 & 0.8 & 2e-5 & 1e-1 & 100 \\
        \cmidrule{2-14}
        & SAN & \textbf{x} & \textbf{x} & \textbf{x} & - & - & - & 7e-4 & 25 & 0.5 & 1e-6 & - & - \\
        & SAN & \textbf{RWPE} & \textbf{\checkmark} & \textbf{x} & 12 & - & - & 7e-4 & 25 & 0.5 & 1e-6 & - & - \\
        \cmidrule{2-14}
        & GraphiT & \textbf{x} & \textbf{x} & \textbf{x} & - & 0.25 & 16 & 7e-4 & 25 & 0.5 & 1e-6 & - & -  \\
        & GraphiT & \textbf{RWPE} & \textbf{\checkmark} & \textbf{x} & 16 & 0.25 & 16 & 7e-4 & 25 & 0.5 & 1e-6 & - & - \\
        \toprule
        \parbox[t]{2mm}{\multirow{8}{*}{\rotatebox[origin=c]{90}{MOLPCBA}}} & \textbf{Model} &  \textbf{Init PE} & \textbf{\texttt{LSPE}} & \textbf{PosLoss} & \textbf{$k$} & \textbf{$\beta$} & \textbf{$p$} & \textbf{Init\_lr} & \textbf{Patience} & \textbf{Factor} & \textbf{Min\_lr} & \textbf{$\alpha$} & \textbf{$\lambda$} \\
        \midrule
        & GatedGCN & \textbf{x} & \textbf{x} & \textbf{x} & - & - & - & 1e-3 & 25 & 0.5 & 1e-4 & - & -\\
        & GatedGCN & \textbf{LapPE} & \textbf{x} & \textbf{x} & 3 & - & - & 1e-3 & 25 & 0.5 & 1e-4 & - & -\\
        & GatedGCN & \textbf{RWPE} & \textbf{\checkmark} & \textbf{x} & 16 & - & - & 1e-3 & 25 & 0.5 & 1e-4 & - & -\\
        \cmidrule{2-14}
        & PNA & \textbf{x} & \textbf{x} & \textbf{x} & - & - & - & 5e-4 & 4 & 0.8 & 2e-5 & - & - \\
        & PNA & \textbf{RWPE} & \textbf{\checkmark} & \textbf{x} & 16 & - & - & 5e-4 & 10 & 0.8 & 2e-5 & - & -\\
        \bottomrule
    \end{tabular}
    }
    \label{tab:model_configurations}
\end{table}

In Table \ref{tab:model_configurations}, additional details on the hyperparameters of different models used in Table \ref{tab:all_lspe_expts} are provided. As for hardware information, all models were trained on Intel Xeon CPU E5-2690 v4 server having 4 Nvidia 1080Ti GPUs, with each single GPU running 1 experiment which equals to 4 parallel experiments on the machine at a single time.

\end{document}